\documentclass[letterpaper, 10 pt, conference]{ieeeconf}  % Comment this line out if you need a4paper

\IEEEoverridecommandlockouts                              % This command is only needed if 
\usepackage{amsmath} % assumes amsmath package installed
\usepackage{hyperref}
\usepackage{amssymb}  % assumes amsmath package installed
\usepackage{dsfont}
\usepackage{graphicx}
\usepackage{subcaption}

\usepackage{psfrag,graphicx,epsfig}
\usepackage{epstopdf}
\usepackage{xspace}
\usepackage{float}
\usepackage{placeins}
\usepackage{multirow}
\usepackage{pgf,tikz}
\usepackage{nowidow}
\usepackage{lineno}
\usepackage{xcolor}
\newcommand{\fig}[1]{Fig.~\ref{#1}}
\newcommand{\tab}[1]{Table~\ref{#1}}
\newcommand{\eq}[1]{(\ref{#1})}

\usepackage{siunitx}
\usepackage{color}
\usepackage{makecell}
\usepackage{lineno}
\hypersetup{
    colorlinks=true,
    linkcolor=black,
    filecolor=magenta,      
    urlcolor=cyan,
    pdfpagemode=FullScreen,
    }
\allowdisplaybreaks

\newcommand{\gurobi}{\textsc{Gurobi}}
\newcommand{\kine}{\textsc{K-Opt}}
\newcommand{\cont}{\textsc{C-Opt}}
\newcommand{\quasi}{\textsc{Q-Opt}}
\begin{document}

\title{\LARGE \bf
Hierarchical Contact-Rich Trajectory Optimization for Multi-Modal Manipulation using Tight Convex Relaxations}

\author{Yuki Shirai, Arvind Raghunathan, and Devesh K. Jha%   
\thanks{Authors are with Mitsubishi Electric Research Laboratories, Cambridge, MA, USA 02139 {\tt\small \{shirai,raghunathan,jha\}@merl.com}}%
% \thanks{$^{\ddagger}$Devesh K. Jha and Arvind U. Raghunathan are with Mitsubishi Electric Research Laboratories (MERL), Cambridge, MA, USA 02139 {\tt\small \{jha,raghunathan\}@merl.com}}
}%

\maketitle

\begin{abstract}
Designing trajectories for manipulation through contact is challenging as it requires reasoning of object \& robot trajectories as well as complex contact sequences simultaneously.
%the reasoning of objects, robots, and contacts simultaneously. 
% This is an especially serious issue as the scale of the problem increases such as time horizon and number of potential contacts. 
In this paper, we present a novel framework for simultaneously designing trajectories of robots, objects, and contacts efficiently for contact-rich manipulation. 
We propose a hierarchical optimization framework where Mixed-Integer Linear Program (MILP) selects optimal contacts between robot \& object using approximate dynamical constraints, and then a NonLinear Program (NLP) optimizes trajectory of the robot(s) and object considering full nonlinear constraints. %This is beneficial as the proposed method utilizes the strength of mixed-integer programming (MIP) and NLP. 
We present a convex relaxation of bilinear constraints using binary encoding technique such that MILP can provide tighter solutions with better computational complexity.
%Furthermore, we employ the cutting-plane method to remove specific contact combinations if NLP is unable to converge. 
The proposed framework is evaluated on various manipulation tasks where it can reason about complex multi-contact interactions while providing computational advantages. We also demonstrate our framework in hardware experiments using a bimanual robot system.  
The video summarizing this paper and hardware experiments is found \href{https://youtu.be/s2S1Eg5RsRE?si=chPkftz_a3NAHxLq}{here}.

% , showing that it can reason about complex, multi-contact interactions 

% design various multi-modal contact-rich trajectories and computationally outperforms the baseline methods. 
% Designing motion planners for contact-rich robotic manipulation is challenging because there are many 
% robotic manipulation is quite challenging planning problem.
% simultaneous method can be fragile
% sequential method but not work well
% make MILP as strong as possible
% if NLP fails, still has something
% evaluated in several cases
\end{abstract}
\section{Introduction}\label{sec:intro}
Humans can effortlessly perform dexterous, multi-modal manipulation tasks while reasoning through complex contact configurations to fully use their whole body during manipulation. 
% For example, in figure, he first rotates his bag, grasps it, lifts it, and then pushes it. 
% \ysnote{put figure?} 
However, achieving similar dexterous behavior for robots is very difficult \cite{ota2024autonomous, billard2019trends, saha2017task}. While there are several critical reasons, we point out a few here. First, planners need to consider long-horizon manipulation to achieve multi-modal, complex behavior \cite{zhao2021sydebo}. This leads to large-scale optimization problems which are difficult to solve. Second, planners need to consider the kinematic, dynamic, and contact constraints of objects and robots simultaneously, resulting in nonlinear, discontinuous constraints that require careful treatment to find meaningful solutions~\cite{toussaint2022sequence, stouraitis2020online}. Considering all possible contact constraints leads to huge computational costs which is undesirable.
Consequently, several manipulation planning algorithms consider fixed contact modes which limits the complexity of tasks which can be solved by these algorithms. Our goal in this paper is to overcome some of these shortcomings by considering reasoning over all contact variables (i.e., locations, sticking, slipping, etc.) while being computationally efficient.
% where all of the contact modes are not fixed but still, the computation is cheap. 
% can be categorized into 1) some contact modes are fixed but others are not, which limits the dexterity of the task, or 2) all contact modes are not fixed but the computation is quite demanding.  

\begin{figure}[t]
        \centering
        \includegraphics[width=0.49\textwidth]{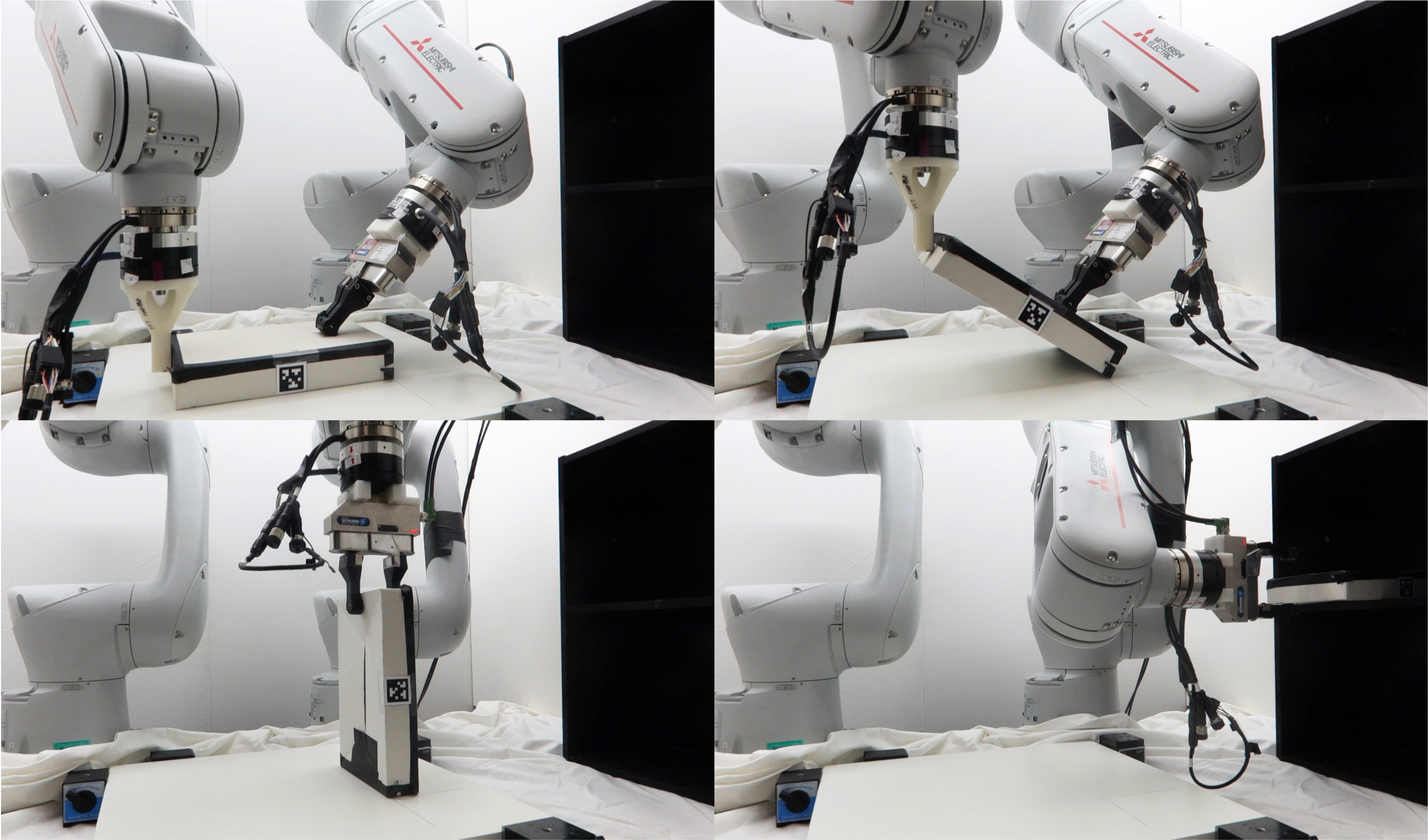}
    \caption{
   We show a bimanual system which can reason about pivoting the box to grasp it so that it can be stowed on a shelf using the proposed algorithm. The Apriltag system is used for feedback during grasping since the box might slip during pivoting.
   The hardware experiment video is found \href{https://youtu.be/s2S1Eg5RsRE?si=g7JP4_0Cchm49c2b}{here}.
    }
        \label{fig1}
\end{figure}

We introduce a hierarchical contact-rich trajectory optimization algorithm that considers kinematics, dynamics, and contact constraints while maintaining computational efficiency. In the proposed hierarchical approach, an MIP plans contact locations and forces between the robot(s) and object while considering approximate constraints. We present a convex relaxation of bilinear constraints, which often appears in cross-product computation of moments and complementarity constraints. This allows our MIP 
formulation to plan trajectories with the tight approximation of nonlinear dynamics. This solution is then passed to an NLP which reasons about the entire manipulation trajectory considering full nonlinear contact constraints. The proposed hierarchical optimization framework can then effectively reason about very complex contact interactions over long planning horizons and can find very rich solutions resulting in multi-modal behavior where a robot can switch between multiple contact points manipulation. This is illustrated through various numerical as well as hardware experiments (see \fig{fig1} for an example of a stowing task).
% We first present a planning algorithm 
% where MIP plans some contact modes (i.e., robot contact location) and the corresponding contact forces, and then NLP considers the trajectory of the object and robot with other contact modes (i.e., sticking-slipping mode). 

% Furthermore, we present feedback cutting plane method, where we remove specific contact combination when NLP fails to converge. 
% 
% The proposed method is extensively tested with various scenarios and some of them are tested in hardware experiments. 

% As a result, the current robotic manipulation planning often tries to avoid changing contact, assumes some contact modes, 
% This is mainly because contacts 

% Contacts are the key to enabling dexterous manipulation because robots can interact with environments through contact \cite{billard2019trends}. 
% 
% In order to fully use manipulation skills, planners considering multi-modality of manipulation is very important.  
% Planning trajectories for manipulation through contact is difficult because planners need to reason complex frictional interactions involving objects, robots, and contacts.
% In general, it is very challenging for planners to design feasible trajectories considering these complex interactions.
% \ysnote{Say robotic manipulation requires the reasoning of object, robot, and contact together. However, it is quite challenging problem. Then say motivation. Let's do iterative optimization that can run faster and more reliably. 
% Also, I want to give more insights. For example, we are interested in book shelf and it is necessary to consider everything. To do this, long-horizon planning is necessary. But it is challenging using SOTA. We have this.}

\textbf{Summary of Contributions.} 
\begin{enumerate}
    \item We present a framework for simultaneously designing trajectories of robots, an object, and contact efficiently with no fixed contact modes. 
    \item We propose a tight and efficient convex relaxation of bilinear constraints using the binary encoding method. %and the method for removing the current contact combination if NLP is unable to converge. 
    \item We extensively verify our approach in multiple manipulation scenarios including hardware experiments. 
\end{enumerate}

% \ysnote{Detail contributions: 
% }

% \begin{enumerate}
%     \item General framework: Using our framework, we can solve MINLP efficiently.
%     \item Tight constraints: Through quadratic approximation, MIQP can find reasonably good solutions.
%     \item Contact Feedback: In order to utilize infeasible case, we do cutting plane. 
%     \item I still want to say something about nested implementation MIQP and NLP. 
%     \item Various experiments 
% \end{enumerate}

\section{Related Work}\label{sec:related_works}
% contact optimization with some works/
% however they still assume some properties
% Also surprisingly there is not so many works which are out of table top, requires more smaller dt, meaning longer horizon
% much less assumptions but does not limit generality and computation is cheap. 

% there are some interesting ones considering bilinear is popular to avoid nonlinear one. BUt not good.
% infeasibility

% 

Contact-Implicit Trajectory Optimization (CITO) has been extensively studied in robotic locomotion and manipulation literature.
These works can be based on  NLP with complementarity constraints \cite{posa2014direct, patel2019contact, moura2022non, zhang2023simultaneous,  shirai2023covariance}, based on MIP \cite{valenzuela2016mixed, aceituno2017simultaneous, aceituno2020global, ding2020kinodynamic, shirai_2022iros}, or based on smoothed dynamics \cite{pang2023global, shirai2024linear}. 
Although these works show impressive results, each work has some assumptions on the contact modes (e.g., Sliding-sticking complementarity contact is not considered in \cite{zhang2023simultaneous, valenzuela2016mixed, aceituno2017simultaneous, shirai_2022iros, graesdal2024convexrelaxations}. Contact patch is fixed in \cite{posa2014direct, moura2022non}. Unphysical artifacts, force-at-distance, are introduced in \cite{pang2023global}.) In \cite{zhang2023simultaneous}, the authors present a general-purpose method to consider contact locations between the object and the environment; however, the proposed method can not consider the selection of contact between the robot and the object. 
% Hence, they might not satisfy our requirements, which are contact-rich trajectories with no fixed contact modes. 
In contrast, the framework proposed in this paper does not have any of the aforementioned assumptions.  %Furthermore, it can also achieve computational efficiency while solving the full problem.

% Another related field is the relaxation of bilinear constraints. Tight relaxation of bilinear constraints is important so MIP can find better-quality solutions. 
% McCormick envelopes approximate bilinear terms with linear estimators, though they can be loose \cite{mccormick1976computability}. Quadratic cost function relaxations \cite{aceituno2017simultaneous, ponton2016convex, ponton2021efficient} also struggle to control bilinear term violations, and tuning them is tricky. 
% % 
% While \cite{valenzuela2016mixed, lin2021designing} consider piecewise McCormick envelopes for tighter solutions, the computation can be still expensive. 
% In this paper, we employ binary encoding \cite{vielma2011modeling}, offering a more efficient approach.

Another related field is the relaxation of bilinear constraints, which is crucial for improving MIP solution quality. McCormick envelopes provide linear approximations of bilinear terms but can be loose \cite{mccormick1976computability}. Quadratic relaxations \cite{aceituno2017simultaneous, ponton2016convex, ponton2021efficient} also struggle to control bilinear violations and require careful tuning.
While piecewise McCormick envelopes offer tighter bounds \cite{valenzuela2016mixed, lin2021designing}, they can be computationally expensive. Instead, we employ binary encoding \cite{vielma2011modeling} for a more efficient approach.

\section{Problem Formulation}\label{sec:problem_statement}
% \ysnote{Is it better to only define variables and assumption here and later explain the constraints in detail?}
We provide an overview of our model and describe the assumptions.
We consider the following assumptions. 
\begin{enumerate}
    \item Rigid objects and robots. 
    \item Quasi-static equilibrium of the system.
    \item Manipulation in SE(2).
    \item Extrinsic contact happens at the vertices of the object.
\end{enumerate}
% It is worth noting that we do not assume tabletop manipulation, change of contact modes (e.g., sticking-sliding), or balance of movement, unlike the work presented in \cite{valenzuela2016mixed, doshi2020hybrid, graesdal2024convexrelaxations}, which incorporate some or all these assumptions.

\begin{figure}[t]
        \centering
        \includegraphics[width=0.25\textwidth]{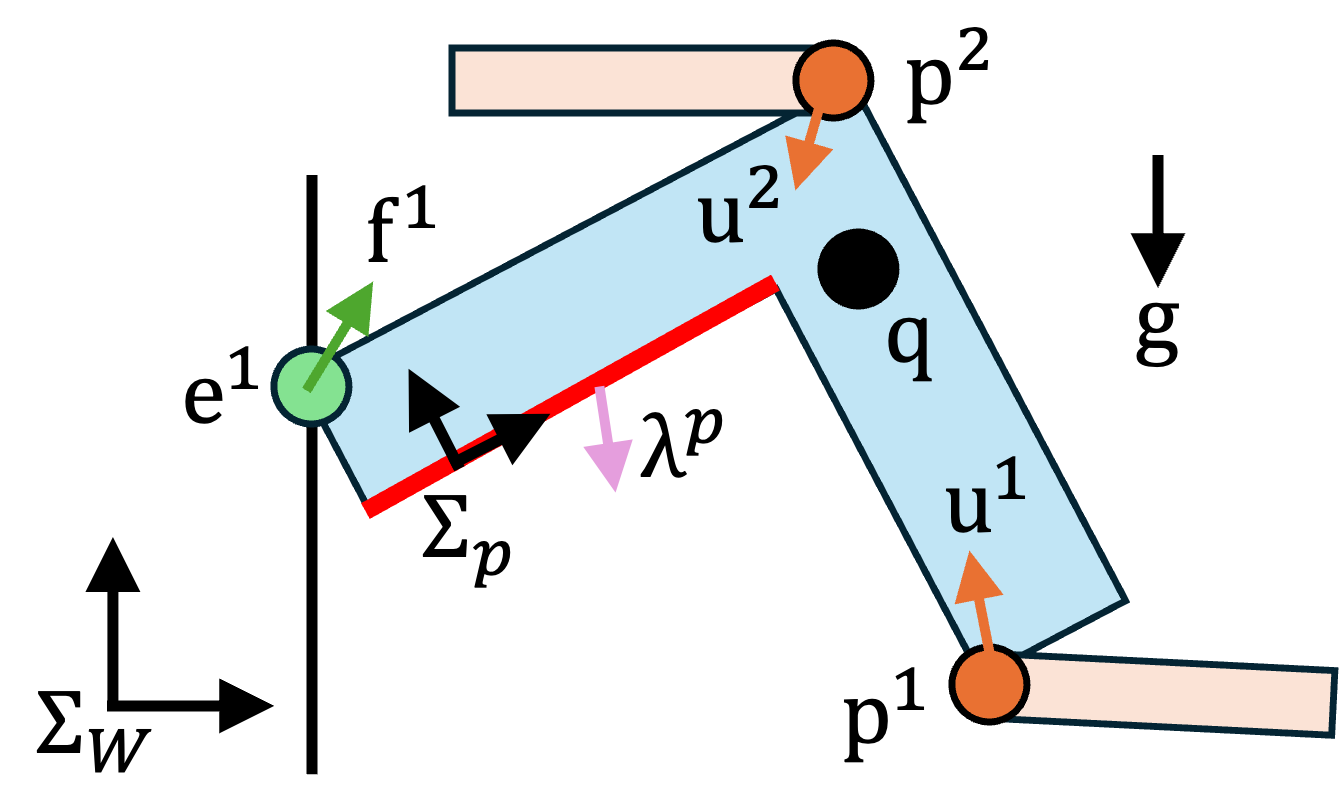}
    \caption{
   A schematic showing the free-body diagram of a rigid body during manipulation where two robots make contact and there is one extrinsic contact between the object and the environment. We consider $N_r = 2$, $N_v = 6$ and $N_p = 6$ in this figure.
   The red line represents one specific object's contact surface with the corresponding local force $\lambda$.
   The variables in this figure are summarized in \tab{tab:my_label}.
    }
        \label{fig:free-body}
\end{figure}

% For extrinsic contacts, we always assume that contacts happen only at the vertices of the object. This is a fair assumption since we can approximate another contact type such as patch contact with two-point contacts, which is often used as generalized friction cones in manipulation planning \cite{chavan2018hand, shirai2023tactile, taylor2023object, shirai2024robust}. 
% Also, this is a weaker assumption than other works \cite{graesdal2024convexrelaxations} assuming tabletop manipulation, which only focuses on discussing the interaction between the robot and the object. 
The free-body diagram of the system we consider in this work is shown in \fig{fig:free-body}.
In \fig{fig:free-body}, it is important to point out that an object can make multiple contacts during manipulation. For example, the object can make contact such as extrinsic contact with the environment and the robot contact.
For extrinsic contacts, we assume that contacts occur only at the object's vertices. This is reasonable since other types, like patch contacts, can be approximated by two-point contacts, commonly used in manipulation planning with generalized friction cones \cite{chavan2018hand, shirai2023tactile, taylor2023object, shirai2024robust}. Additionally, this assumption is less restrictive than those in works like \cite{graesdal2024convexrelaxations}, which focus solely on tabletop manipulation and the interaction between the robot and the object.

\begin{table}[]
    \centering
        \caption{Notation of variables. C or B indicates the variable is a continuous or binary variable, respectively. In $\Sigma$ column, we indicate the frame of variables. Subscript $t$ indicates time-step.}
\begin{tabular}{|c|c|c|c|c|}
\hline Name & Description & Size  & C/B & $\Sigma$ \\
\hline $\mathbf{q}_t$ & object pose & $\mathbb{R}^3$ & C & $W$ \\
$\mathbf{p}_t^i$ & $i$-th robot position & $\mathbb{R}^2$ & C & $W$ \\
$\mathbf{e}_t^v$ & $v$-th extrinsic contact position & $\mathbb{R}^{2}$ & C & $W$ \\
$\mathbf{u}_t^i$ & $i$-th robot contact force & $\mathbb{R}^2$ & C & $W$ \\
 $\boldsymbol{\lambda}^{i, p}_t$ & $p$-th local force for $i$-th robot & $\mathbb{R}^2$ & C & $p$ \\
 $\mathbf{f}^{v}_t$ & $v$-th extrinsic contact force & $\mathbb{R}^2$ & C & $W$ \\
\hline $z^{i, p}_t$ & $i$-th robot contact at $\mathcal{P}_p$ & $\mathbb{Z}^1$ & B &  \\
\hline
\end{tabular}
    \label{tab:my_label}
\end{table}

The variables used in this paper are summarized in \tab{tab:my_label} and \fig{fig:free-body}.
In this work, we consider in total $N_r$ robots and a single object which consists of $N_v$ potential extrinsic contacts and $N_p$ potential object surfaces where the robot can make contact.
We denote $\mathcal{P}_p$ as the $p$-th object surface, associated with a local frame $\Sigma_{p}$.
$\mathcal{P}_p$ is represented as halfspace.
For any arbitrary vector $\mathbf{x}$, the notation $\|\mathbf{x}\|^2_{Q}$ means a quadratic term with a positive-semi-definite matrix $Q$.
We define the coordinate transformation and rotation matrix from frame $\Sigma_A$ to $\Sigma_B$ as $^A_BT$ and $^A_BR$, respectively.
We denote $X \Longrightarrow Y$ as a conditional constraint and implement it using a big-M formulation in MIP \cite{balas1979disjunctive}. 

% For each time-step $t$, $\mathbf{p}_t^i \in \mathbb{R}^2$ and $\dot{\mathbf{p}}_t^i \in \mathbb{R}^2$ represents the location of the $i$-th robot and its derivative. $\mathbf{e}_t^j \in \mathbb{R}^2$ and $\dot{\mathbf{e}}_t^j \in \mathbb{R}^2$ is the extrinsic contact location and its derivative. $\mathbf{u}_t^i \in \mathbb{R}^2$ represents the contact force from the $i$-th robot to the object. $\mathbf{f}_t^i \in \mathbb{R}^2$ represents the extrinsic contact force. $\mathbf{q}_t \in \mathbb{SE}(2)$ and $\dot{\mathbf{q}}_t \in \mathbb{SE}(2)$ is the pose of the object and its derivative. 
The constants $m$ and $\mathbf{g} \in \mathbb{R}^2$ represent the mass of the object and gravitational acceleration, defined in world frame $\Sigma_W$, respectively.  $\mu_e^v$ and $\mu_r^i$ are the friction constraints for $v$-th extrinsic contact and $i$-th robot contact, respectively.
$h$ is the step size.
We use subscripts $n$ and $s$ to represent normal and tangential elements of forces.
% We use subscripts $+$ and $-$ to represent positive and negative direction of the velocity. 
Also, we use subscripts $x, y, \theta$, to represent the element of $\mathbf{q}_t$.
% For brevity, we omit subscript $t$ for the following discussion in this section. 

% \subsection{}

% \subsection{Contact-Implicit Trajectory Optimization}
% We formulate the following optimal control problem.
% \begin{subequations}
% \begin{flalign}
% % \min _{x, u, \lambda}\sum_{k=0}^{N-1} \phi(x_k, u_k, \lambda_k)\\
% \min _{x, u, f}  \sum_{k=1}^{N} ({x}_{k} - x_g)^{\top} Q ({x}_{k} - x_g)+\sum_{k=0}^{N-1}u_{k}^{\top} R u_{k} \\
% \text{s. t. } i_{k, x}, i_{k, y} \in FK(\theta_k, \revise{P}_{k, y}^{\revise{O}}), ,  \label{const2}\\
% x_{0} = x_s, x_{N} = x_g,
% x_{k} \in \mathcal{X}, u_{k} \in \mathcal{U}, 0\leq f_{k, ni} \leq f_{u} \label{bounds_variables}
% \end{flalign}
% \label{equation_control}
% \end{subequations}

% As shown in, this optimization problem would be MINLP, which is quite difficult to find solution in general

% \begin{subequations}
% \begin{flalign}
% \alpha_t^{i, c} = 1 &\Longrightarrow
% \left \{
% \begin{aligned}
% \mathbf{f}_t^{i, c}, \mathbf{m}_t^{i, c} \in \mathcal{W}\left(\mu_c, k_c\right),\\ \dot{\mathbf{p}}_t^i, \dot{\mathbf{q}}_t^i  = \mathbf{0}, 
% {^{c}_WT} \left(\mathbf{p}_t^i, \mathbf{q}_t^i\right) \in \mathcal{C}_c
% \end{aligned}
% \right\}\label{contactA}
% \\
% \alpha_t^{i, c} = 0 &\Longrightarrow \mathbf{f}_t^{i, c}, \mathbf{m}_t^{i, c} = \mathbf{0}\label{contactB}\\
% \sum_{c=1}^{C} \alpha_t^{i, c} &\leq 1 \label{sum_contact}
% \end{flalign}
% \label{contact_eq}
% \end{subequations}
\section{Contact-Rich Trajectory Optimization}\label{sec:main_algorithm}
\begin{figure}[t]
    
        \centering
        \includegraphics[width=0.4\textwidth]{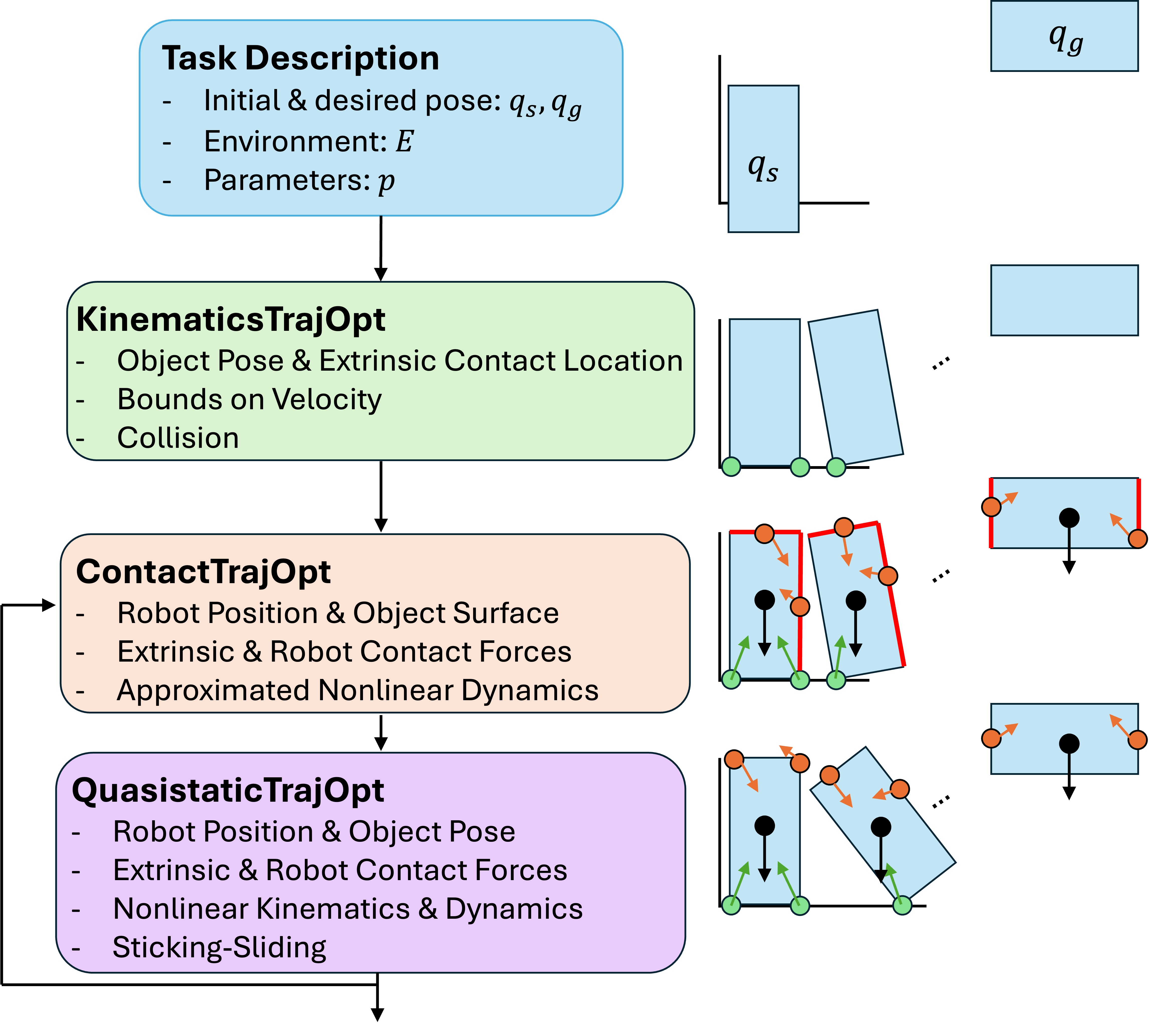}

    \caption{
   Overview of the proposed framework. See Sec~\ref{overview_sec} for details.  Orange points and green points represent the robot contact location and extrinsic contact location, respectively. 
   Red lines represent the object's contact surface where the robot makes contact. 
   We have the cutting plane method to deal with the infeasible solution by \quasi{} (see Sec~\ref{sec:feedback_cutting}).
   % Given the task description, we first run \kine{}, which returns kinematically feasible trajectories of the object and extrinsic contact location. 
   % We then run \cont{}, which returns trajectories of the extrinsic contact forces as well as robot positions and the robot contact forces using approximated dynamics. 
   % Finally, we run \quasi{}, which returns trajectories of object pose, robot position, and extrinsic and robot contact forces using the actual nonlinear dynamics with fixed contact patches. 
   % For each module, we show the frequency that optimizers can run. These numbers are measured when we consider the pivoting for the box object with $T=150$. 
    }
            \label{fig:overview_method}
\end{figure}

A naive formulation of trajectory optimization using all possible contact constraints would result in a mixed-integer nonlinear program, which could be computationally very demanding to solve. We propose a method that solves this problem using hierarchical optimization (see \fig{fig:overview_method}). 
% The details of the proposed method are provided in this section.
% A high-level overview is shown in \fig{fig:overview_method}.

% If we naively formulate trajectory optimization problem using the constraints presented in this section, the resulting optimization problem becomes mixed-integer nonlinear program (MINLP), which is in general very difficult to find solutions. 
% 
% 
% In this section, we first present the overview of our proposed method. Second, we describe some key contributions of this method. Finally, we explain the constraints used in this method. 
% \ysnote{explain why making-breaking is in MILP and stick-slip is in NLP}

\subsection{Overview}\label{overview_sec}
Our framework is illustrated in \fig{fig:overview_method}. Our framework consists of three optimization problems. First, given the task description, Kinematics Trajectory Optimization, \kine{}, finds a feasible trajectory of the pose of the object and extrinsic contacts while satisfying collision constraints between the object and environment. Second, given the solution by \kine{}, Contact Trajectory Optimization, \cont{}, finds a feasible trajectory of extrinsic contact forces, the robot position, and the robot contact force on the contact location between the robot and the object.
 % and the corresponding robot contact states as well as the robot contact forces while satisfying approximated dynamics constraints. 
 Finally, using this solution, Quasistatic Trajectory Optimization, \quasi{}, finds a feasible trajectory of the pose of the object, robot position, and the extrinsic \& robot contact forces while satisfying full nonlinear dynamic and kinematic constraints. In \quasi{}, the object surface where the robot makes contact is fixed from \cont{} but sliding-sticking complementarity constraints are still considered. Therefore, our method has the flexibility to adjust the robot position such that it can find much richer solutions resulting in more dexterous behavior. 
It is possible that \quasi{} is unable to find a feasible solution depending on the solution returned by MIP (since MIP considers approximate constraints). In such cases, we remove the specific contact combination from \cont{} and re-run \cont{} based on the cutting-plane method, which is explained in Sec~\ref{sec:feedback_cutting}. 
% 
% We explain each of the modules in detail.

\subsubsection{Kinematics Trajectory Optimization}\label{sec:kine_trajopt}
In \kine{}, the goal is to find feasible trajectories of the pose of the object and extrinsic contacts between an object and the environment. We consider the following optimization problem.
\begin{subequations}
\begin{flalign}
% \min _{x, u, \lambda}\sum_{k=0}^{N-1} \phi(x_k, u_k, \lambda_k)\\
\min _{\mathbf{q}_t,  \dot{\mathbf{q}}_{t}}  \sum_{t=0}^{T} \|\mathbf{q}_{t} - \mathbf{q}_t^\text{ref}\|^2_{Q_\text{kin}}\\
\text{s. t. } ,  \mathbf{q}_{t+1} = \mathbf{q}_{t} + h \dot{\mathbf{q}}_{t} \label{const2}\\
% \mathbf{q}_{0}  = \mathbf{q}_{s}\\
\underline{\mathbf{q}}_t\leq\mathbf{q}_t \leq \overline{\mathbf{q}}_t, \underline{\dot{\mathbf{q}}}_t\leq\dot{\mathbf{q}}_t \leq \overline{\dot{\mathbf{q}}}_t, \label{bound_q}\\
\text{sdf}(\mathbf{q}_t)\geq 0, \label{sdf_eq}
\end{flalign}
\label{kine_eq}
\end{subequations}
where $\mathbf{q}_t^\text{ref}$ is the linear interpolation between $\mathbf{q}_\text{s}$ and $\mathbf{q}_\text{g}$ with $T$ steps, and $h$ is time interval. 
% $h \geq 0$ is the time-interval for discretization. 
\eq{const2} is the dynamics of object pose and \eq{bound_q} is the bound of variables. 
sdf is the signed distance function between the object and the environment which computes the distance between $e_t^v$ and the environment. 
None of the constraints in \eq{kine_eq} involve any integer variables and thus \eq{kine_eq} is formulated as NLP. 

After solving \eq{kine_eq}, we compute a binary map $A$, which tells if each extrinsic contact of the object, $\mathbf{e}_t^v$, makes contact with the environment, where $\mathbf{e}_t^v$ is calculated as a function of $\mathbf{q}_t$ and the object \& environment geometry.
Hence, \cont{} does not need to reason about $\mathbf{e}_t^v$.
Similarly, we compute a binary map $B$ to tell which of the extrinsic contacts slip. This helps us consider the correct friction cone constraints for downstream optimization.
% , where $'1'$ indicates the contact slips.
We denote $x_\text{kin}:=[\mathbf{q}_t, A, B, \forall t]$ to represent the solution from this module. 

% After solving \eq{kine_eq}, we can compute if each extrinsic contact of the object, $\mathbf{e}_t^v$, makes contact with the environment and introduce a binary map, $A \in \{0, 1\}^{\{N_e \times T\}}$, where $'1'$ indicates the contact is made.
% Similarly, we can compute if each extrinsic contact of the object slips or not and introduce another binary map, $B \in \{0, 1\}^{\{N_e \times T-1\}}$.
% % , where $'1'$ indicates the contact slips.
% We denote $x_\text{kin}:=[\mathbf{q}_t, A, B, \forall t]$ to represent the solution from this module. 

% It is worth noting that we already start observing some benefits of the proposed method. 
% Using \kine{}, \cont{} does not need to reason about extrinsic contact states between the object and the environment, which decreases the number of integer constraints.
% introduce a number of integer constraints or complementarity constraints. Also, we can get 
\subsubsection{Contact Trajectory Optimization}
In \cont{}, the objective is to find extrinsic contact forces, contact location of the robot, object surfaces where the robot makes contact, and contact forces from the robot for a fixed $x_\text{kin}$. 
The optimization problem is then formulated as:
\begin{subequations}
\begin{flalign}
% \min _{x, u, \lambda}\sum_{k=0}^{N-1} \phi(x_k, u_k, \lambda_k)\\
\text{Find} ~ \mathbf{p}_t^i, \mathbf{u}_t^i, \mathbf{f}_t^v, \boldsymbol{\lambda}^{i, p}_t, {z}_t^{i, p}, ~ \forall t, i, v, p \\
\text{s. t. },   \mathbf{p}_{t+1} = \mathbf{p}_{t} + h \dot{\mathbf{p}}_{t} \label{finger_kinematics} \\
\underline{\dot{\mathbf{p}}}_t^i \leq\dot{\mathbf{p}}_t^i \leq \overline{\dot{\mathbf{p}}^i_t}, \label{finger_bound}\\
\mathbf{u}_t^i   = \sum_{p=1}^{N_p}{^W_pR}\boldsymbol{\lambda}^{i, p}_t \label{convert_force}\\
\mathbf{p}_{t}^i \in \mathcal{Q}_\text{surface} \cap \mathcal{Q}_\text{mom}  \label{cont_patch} \\
\boldsymbol{\lambda}^{i, p}_t \in 
\mathcal{Q}_\text{stable} 
\cap 
\mathcal{Q}_\text{surface} 
\cap 
\mathcal{Q}_\text{force} \cap \mathcal{Q}_\text{mom}\cap \mathcal{Q}_\text{FC} \cap  \mathcal{Q}_\text{C}  \label{cont_lambda}
\\
\mathbf{f}_t^v \in \mathcal{Q}_\text{force} \cap \mathcal{Q}_\text{mom} \cap \mathcal{Q}_\text{FC} \label{cont_f}\\
{z}_t^{i, p} \in  \mathcal{Q}_\text{surface} \cap \in 
\mathcal{Q}_\text{stable}   \cap  \mathcal{Q}_\text{C} 
% \\
% \eq{Mechanicsforceeq1}, \eq{contact_eq}, \eq{Mechanicsmoment_eq1}, \text{Change of Contact Constraints}
\end{flalign}
\label{cont_eq}
\end{subequations}
where $\mathcal{Q}_\text{surface},\mathcal{Q}_\text{force},\mathcal{Q}_\text{mom}, \mathcal{Q}_\text{stable}, \mathcal{Q}_\text{FC}$, and $\mathcal{Q}_\text{C}$ are the set encoding object surface selection constraint, force balance constraint, moment balance constraint,  stable contact change constraint,  friction cone constant, and integer constraint, respectively.
We explain each set in Sec.~\ref{const_sec}.
\eq{finger_kinematics} is the dynamics of the robot position and \eq{finger_bound} considers bound of robot velocity. With \eq{making_eq}, \eq{convert_force} converts a  specific local wrench in $\Sigma_p$ to the wrench in $\Sigma_W$.
\eq{cont_eq} is a feasibility problem. 
% \eq{convert_force} is 

In \eq{cont_eq}, the problem is formulated as mixed-integer \textit{non-convex} Quadratically Constrained QP (QCQP) due to the bilinear term in $\mathcal{Q}_\text{mom}$, which is in general quite tough to find a feasible solution.
Thus, we propose a convex relaxation of bilinear terms such that \eq{cont_eq} is formulated as MILP, which is much easier to find a feasible solution while improving the computational complexity. 
% Also, thanks to \kine{}, we do not introduce any integer variables for extrinsic contact since we know the extrinsic contact pattern using the map $A$. 
% Similarly, we also know if the extrinsic contact slips from the map $B$. In practice, we observe that these maps dramatically decrease the search space of the optimization problem and \cont{} runs much more stably. 
% $z_t^{i, p}$
% 
After solving \cont{}, \cont{} returns $x_\text{con}:=[\mathbf{p}_t^i, \mathbf{u}_t^i, \mathbf{f}_t^v, \boldsymbol{\lambda}^{i, p}_t, {z}_t^{i, p},  \forall t, i, v, p ]$.

In \cont{}, we do not consider sticking-sliding complementarity constraints. This is mainly for computational reasons. If we consider these constraints in \cont{}, we have many more complicated integer constraints, which makes \cont{} slow. Instead, we propose to let \quasi{} consider sticking-sliding complementarity constraints while making pose of the object, $\mathbf{q}$ as a decision variable. As a result, we can achieve better computation. 

% We argue that \cont{} shows quite good performance. It can find a globally optimal solution. Also, even though this is MIQCQP, the QCQP is now convex, which can reliably run. And the all of these benefits appear since we decompose the original optimization problem such that \cont{}focuses on only making-breaking constraints as a source of integer constraints.

% In \eq{cont_eq}, \eq{Mechanicsforceeq1} is actually linear constraints since $\mathbf{q}$ is fixed from $x_\text{kin}$. \eq{contact_eq} is mixed-integer linear constraints, which are formulated using big-M formulation. \eq{Mechanicsmoment_eq1} is nonlinear constraints due to the bilinear terms. Change of Contact Constraints are formulated such as 

% Therefore, \eq{cont_eq} is mixed-integer non-convex QCQP problem, which is in general tough for the solver to converge. The most naive way to deal with this issue is that we simply remove \eq{Mechanicsmoment_eq1}, which can result in a quite bad solution. As a result, \quasi{} will have difficulties finding solutions since there is no solution

% \cite{zhang2023simultaneous} will take time to find physically feasible transition since it does not consider change of contact modes in their formulation. In contrast, thanks to this constraint, our method can efficiently find the transition of the contact. 

% The advantage of \cont{} is that it only needs to 

% The advantages of this formulation compared to the formulation that has all of constraints in \kine{} and \cont{} are:
% \begin{enumerate}
%     \item The scale of the optimization is smaller. 
% \end{enumerate}

\textit{Remark.} It is possible that \kine{} does not provide a physically feasible $x_\text{kin}$ and thus \cont{} finds an infeasible solution. If this happens, we make ${\mathbf{q}}_{t}$ as also part of the decision variable of \cont{}.
% Rest of the details are skipped for brevity.
% We often get started only making the position of the object decision variables since it is often enough so that we can avoid making a rotation matrix as part of decision variables. 
% As a result, we have more bilinear terms in other constraint sets, which can also be approximated using our relaxation in Sec~\ref{sec:relax_bilinear}.
% It turns out that we have bilinear terms in $\mathcal{Q}_\text{patch},\mathcal{Q}_\text{force}$ as well 
% In practice, we observe that the solution from \cont{} is still decent. If it is not decent, \quasi{} will not able to converge and remove that specific contact combination from \cont{}, which improves the solution quality of \cont{}.

\subsubsection{Quasi-static Trajectory Optimization}
The objective of \quasi{} is to find optimal trajectories of extrinsic contact forces and robot forces, robot position, and object pose while fixing $z^{i, p}_t$ (i.e., red lines in \fig{fig:overview_method}) and $A$.
\begin{subequations}
\begin{flalign}
\min _{\mathbf{q}_t,  \dot{\mathbf{q}}_{t}}  \sum_{t=0}^{T} \|\mathbf{q}_{t} - \mathbf{q}_t^\text{ref}\|^2_{Q_\text{kin}}\\
\text{s. t. } ,  \eq{const2}, \eq{bound_q},  \eq{finger_kinematics}, \eq{finger_bound}, \eq{convert_force}, \eq{cont_patch}, \eq{cont_lambda}, \eq{cont_f}, \\
\boldsymbol{\lambda}^{i, p}_t  \in \mathcal{Q}_{\text{slip}}, \mathbf{f}_t^v   \in \mathcal{Q}_{\text{slip}}, \dot{\mathbf{p}}_t^i  \in \mathcal{Q}_{\text{slip}}, \dot{\mathbf{e}}_t^v \in \mathcal{Q}_{\text{slip}}
% \\
% \mathbf{e}_t^v \in \mathcal{Q}_\text{env}, 
\end{flalign}
\label{quasi_eq}
\end{subequations}
where $\mathcal{Q}_{\text{slip}}$ encodes sticking-slipping complementarity constraints. 
% where $\mathcal{Q}_{\text{slip}}$ and $\mathcal{Q}_{\text{env}}$ are the set encoding sticking-slipping complementarity constraints and the constraint ensuring that extrinsic contact makes contact with the environment, respectively. 
% \quasi{} allows sticking-slipping complementarity constraints at object contact 
Since the robot contact surface and which extrinsic contact makes contact with the environment is fixed, \quasi{} becomes NLP, which can be solved quickly.
% In practice, we use SNOPT and the solution status is always '1'  \cite{gill2018user}, which indicates that the solution is found.

\subsection{Convex Relaxation of Bilinear Terms}\label{sec:relax_bilinear}

We propose a novel approximation of bilinear terms, which appear in \eq{Mechanicsmoment_eq1}, inspired by \cite{vielma2011modeling}.
The key idea is that we want to achieve a tighter approximation of bilinear terms than the naive McCormick envelope relaxation but do not want to increase the computation. 
% The key idea is to run the optimizer fast enough. 
We consider the following bilinear constraint $xy$, where $x\in \{x^L, x^U\}$ and $y\in \{y^L, y^U\}$. 
Then, McCormick envelope is given by: 
% For actual implementation, we normalize the decision variables.
% Using the McCormick envelope, we consider the following relaxation of bilinear constraints
% \begin{equation}
%    w\geq 0, w\geq x + y - 1,
% w\leq x, 
% w\leq y
% \label{McCormick} 
% \end{equation}
% \begin{subequations}
% \begin{flalign}
%   w \geq x^L y + y^L x-x^L y^L, 
%   w \geq x^U y + y^U x-x^U y^U, \\
%   w \leq x^U y + y^L x-x^U y^L, 
%   w \leq x^L y + y^U x-x^L y^U, 
%  % \sum_{c=1}^C \frac{j_c}{C_y} z_c \leq b\leq \sum_{c=1}^C \frac{j_c+1}{C_y} z_c  
% \end{flalign}
% \label{McCormick}
% \end{subequations}
\begin{equation}
   \mathcal{W}= \left\{ 
   \begin{aligned}
   &w, x, y, \\ &x^L, x^U, y^L, y^U 
   \end{aligned}
   \middle| \ 
    \begin{aligned}
        &  w \geq x^L y + y^L x-x^L y^L\\  
        &w \geq x^U y + y^U x-x^U y^U\\
        & w \leq x^U y + y^L x-x^U y^L\\
        &  w \leq x^L y + y^U x-x^L y^U 
    \end{aligned} 
    \right\}
    \label{McCormick}
\end{equation}
where $w$ is used to represent $xy$.

We can obtain a tighter relaxation by partitioning the domain. Here, we consider $C$-regions by introducing binary variables, $\eta_c, c=1, \cdots, C$ for each partition. 
Then, the partitioned McCormick envelopes are given by \cite{valenzuela2016mixed}:
\begin{subequations}
\begin{flalign}
% \min _{x, u, \lambda}\sum_{k=0}^{N-1} \phi(x_k, u_k, \lambda_k)\\
 \sum_{c=1}^C{\eta_c} = 1, \\
 \{{\eta}_c = 1\}    \Longrightarrow \mathcal{W} (w, x, y, x_{c}^L, x_{c}^U, y_{c}^L, y_{c}^U), \\
 \{{\eta}_c = 1\}    \Longrightarrow 
 x_{c}^L \leq x\leq x_{c}^U, y_{c}^L \leq y\leq y_{c}^U 
 % \sum_{c=1}^C x_{L,c}z_c \leq x\leq \sum_{c=1}^C x_{U,c}z_c,   \\
  % \sum_{c=1}^C y_{L,c}z_c \leq y\leq \sum_{c=1}^C y_{U,c}z_c,  \\
  % w \geq x_{c}^L y + y_{c}^L x-x_{c}^L y_{c}^L, 
  % w \geq x_{c}^U y + y_{c}^U x-x_{c}^U y_{c}^U, \\
  % w \leq x_{c}^U y + y_{c}^L x-x_{c}^U y_{c}^L, 
  % w \leq x_{c}^L y + y_{c}^U x-x_{c}^L y_{c}^U, 
 % \sum_{c=1}^C \frac{j_c}{C_y} z_c \leq b\leq \sum_{c=1}^C \frac{j_c+1}{C_y} z_c  
\end{flalign}
\label{general_relaxation}
\end{subequations}
% where $w_c$ is the estimate of bilinear term in $c$-th region and subscripts $(L,c)$ and $(U,c)$ represent lower- and upper-bound of decision variable at $c$-th region, respectively.
However, this approach requires $O(C)$ binary variables, leading to scalability issues for large $C$.
% the number of binary variables increases linearly in \eq{general_relaxation}, which can result in large computations. 

We present the binary encoding technique, which assigns each option a unique binary code, reducing the number of binary variables from $O(C)$ to $O(\log{C})$.
% For instance, if $C=4$, using the method in \eq{general_relaxation}, you introduce $4$ binary variables. However 
Let $K := \log_2C$ and binary variable, $\nu_k, k = 1, \cdots, K$.
Each region $c$ corresponds to a unique binary code, $(d_{c1}, d_{c2}, \cdots, d_{cK})$, where $d_{ci} \in \{0, 1\}$. $d_{ci}$ is the pre-computed $i$-th bit of the binary code for region $c$.
Note that $d_{ci}$ is not a decision variable.
For example, consider $C=4$ regions, indicating $K=2$ binary variables where binary codes are $(0, 0), (0, 1), (1, 0), (1, 1)$. For the region $(1, 0)$, $d_{c1} = 1$ and $d_{c2} = 0$.
We also introduce auxiliary \textit{continuous} decision variables $s_{ci} \in [0, 1]$ for each region.
% Then, using bi
% We consider $c_x$ and $c_y$ intervals for $x$ and $y$, respectively, with $c_x + c_y=K$. 
% For simplicity, here we show the specific case, $c_x = 4, c_y=2$. In this case, 
The resulting formulation is given by.
\begin{subequations}
\begin{flalign}
s_{ck}\geq | \nu_k - d_{ck}|,
% s_{ck}\geq d_{ck}-\nu_k, \label{s_lb}\\
s_{ck}\leq M(\nu_k+d_{ck}-2\nu_kd_{ck}), \label{s_ub}\\
\sum_{k=1}^K s_{ck} = \eta_c  \\
 \{{\eta}_c = 0\}    \Longrightarrow \mathcal{W} (w, x, y, x_{c}^L, x_{c}^U, y_{c}^L, y_{c}^U), \label{eq_mc_piece_encoding} \\
 \{{\eta}_c = 0\}    \Longrightarrow 
 x_{c}^L \leq x\leq x_{c}^U, y_{c}^L \leq y\leq y_{c}^U \label{eq_bound_encoding} 
 % \sum_{c=1}^C x_{L,c}z_c \leq x\leq \sum_{c=1}^C x_{U,c}z_c,   \\
  % \sum_{c=1}^C y_{L,c}z_c \leq y\leq \sum_{c=1}^C y_{U,c}z_c,  \\
  % w \geq x_{c}^L y + y_{c}^L x-x_{c}^L y_{c}^L, 
  % w \geq x_{c}^U y + y_{c}^U x-x_{c}^U y_{c}^U, \\
  % w \leq x_{c}^U y + y_{c}^L x-x_{c}^U y_{c}^L, 
  % w \leq x_{c}^L y + y_{c}^U x-x_{c}^L y_{c}^U, 
 % \sum_{c=1}^C \frac{j_c}{C_y} z_c \leq b\leq \sum_{c=1}^C \frac{j_c+1}{C_y} z_c  
\end{flalign}
\label{binary_encode}
\end{subequations}
where $M$ is a large positive number. 
We emphasize that $\eta$ used in \eq{binary_encode} is \textit{continuous} variable bound in $[0, 1]$ and is different from $\eta$ in \eq{general_relaxation} where $\eta$ is \textit{binary} variable. 
The idea of \eq{binary_encode} is to activate piecewise McCormick envelope constraints \eq{eq_mc_piece_encoding} and the corresponding bound constraints for $x$ and $y$ \eq{eq_bound_encoding}, $s_c$ is enforced to take binary values although it is a continuous variable. This behavior is achieved because:
\begin{itemize}
    \item For $c$-th region, $s_{ck} =0, \forall k$ if $c$-th region is active and $s_{ck}=1$ otherwise from \eq{s_ub} and $s_{ci} \in [0, 1]$.
    \item $\eta_c = 0$ if  $c$-th region is active and $s_c\geq1$ otherwise. 
    \item Only one region is active since the binary encoding corresponds to one region for any $\nu$.
\end{itemize}
Therefore, our MILP formulation \eq{binary_encode} achieves the same tightness of the naive method in \eq{general_relaxation} with the logarithmic number of binary variables.

\subsection{Feedback Cutting Plane with Infeasible Solutions}\label{sec:feedback_cutting}
% It is possible that \cont{} might provide an unphysical solution and thus \quasi{} might only find infeasible solutions. 

Many hierarchical optimization methods do not address cases where solvers find infeasible solutions \cite{wang2019manipulation, sleiman2023versatile, rigo2024hierarchical}.
As some works can deal with infeasible solutions \cite{lin2024accelerate, pan2014predicting}, our method can also find and handle infeasible solutions. 
If \quasi{} only finds infeasible solutions, we remove that specific contact combination from \cont{} and re-run \cont{} and \quasi{}. As a consequence, we observe that \quasi{} has a much higher chance of finding feasible solutions. 

As explained in Sec~\ref{sec_making_breaking}, we consider that $i$-th robot makes contact at object surface $p$ at $t$ if $z_t^{i, p} = 1$ and the robot does not make contact otherwise. To achieve this:
\begin{equation}
        \sum_{t\in \mathcal{T}, i \in \mathcal{I}, p \in \mathcal{P}} z_t^{i, p} \leq N -1
    % \prod_{t\in \mathcal{T}} \prod_{i \in \mathcal{I}} \prod_{p \in \mathcal{P}} z_t^{i, p} =0
    \label{feedback_eq_zero}
\end{equation}
where $\mathcal{T}, \mathcal{I}, \mathcal{P}$ indicate the set of indices where $z_t^{i, p} = 1$ and $N$ is the total number of indices with $z_t^{i, p} = 1$.
\eq{feedback_eq_zero} means that at least one of $z_t^{i, p} = 1$ in $\mathcal{T}, \mathcal{I}, \mathcal{P}$ needs to be zero. 
% In practice, directly implementing \eq{feedback_eq_zero} is computationally demanding and thus we consider the following constraint.
% \begin{equation}
%     \sum_{t\in \mathcal{T}, i \in \mathcal{I}, p \in \mathcal{P}} z_t^{i, p} \leq N -1
% \end{equation}
% where $N$ is the total number of indices with $z_t^{i, p} = 1$.

\subsection{Constraints}\label{const_sec}
% In this section, we describe our constraints.
For brevity, we omit subscript $t$ unless necessary. 

\subsubsection{Quasi-Static Equilibrium}
We consider the quasi-static equilibrium as follows. 
\begin{subequations}
\begin{flalign}
\mathcal{Q}_\text{force} = \{ \mathbf{f}^v, \boldsymbol{\lambda}^{i, p} | F(\mathbf{f}^v, \boldsymbol{\lambda}^{i, p}, \mathbf{q}, m\mathbf{g})  = \mathbf{0}\},\label{Mechanicsforceeq1}\\
\mathcal{Q}_\text{mom} = \{ \mathbf{f}^v, \boldsymbol{\lambda}^{i, p}, \mathbf{p}^i  | G(\mathbf{f}_t^v, \boldsymbol{\lambda}^{i, p}, \mathbf{q}, m\mathbf{g}, \mathbf{p}^i, \mathbf{e}^v) = 0 \}\label{Mechanicsmoment_eq1}
\end{flalign}
\label{Mechanics}
\end{subequations}
where $F$ and $G$ represent static equilibrium of force and moment, respectively. 
In particular, $G$ has multiple bilinear terms due to cross product of contact location and forces.
See \cite{shirai2024robust, hou2020manipulation} for more details.
% Note that \eq{Mechanics} has $\mathbf{q}$ to compute transformation between $\Sigma_W$ and $\Sigma_p$.

% \subsubsection{Contact Model}\label{contact_model_sec}
% \textit{Remark}. 
% For extrinsic contacts, we always assume that contacts happen only at the vertices of the object. This is a fair assumption since we can approximate another contact type such as patch contact with two-point contacts, which is often used as generalized friction cones in manipulation planning \cite{chavan2018hand, shirai2023tactile, taylor2023object, shirai2024robust}. 
% Also, this is a weaker assumption than other works \cite{graesdal2024convexrelaxations} assuming tabletop manipulation, which only focuses on discussing interaction between the robot and the object. 
% In contrast, we let robots make contact anywhere in the objects. 

\subsubsection{Making-Breaking Contact}\label{sec_making_breaking}
% We model making-breaking contact here. 
We consider the following hybrid contact models for each robot contact, not for extrinsic contact between the object and the environment since it is encoded in \kine{} (see Sec~\ref{sec:kine_trajopt}).
% For each robot contact, we consider the following hybrid contact models. Note that we do not need to consider making-breaking constraints for extrinsic contact between the object and the environment since it is encoded in \kine{} (see Sec~\ref{sec:kine_trajopt}).
% Given binary variable $z_e^j \in [0, 1]$
% \begin{subequations}
% \begin{flalign}
% \{{z}^{i, p} = 1\} &\Longrightarrow
% \left \{
% \begin{aligned}
% \lambda_n \geq 0, |\lambda_f|\leq \mu \lambda_n\\  
%  y_n = 0, |y_f| \leq y_\text{bound}
% \end{aligned}
% \right\}\label{contactA}
% \\
% \{{z}^{i, p} = 0\}   &\Longrightarrow {\lambda}^{i, p}_n = {0}\label{contactB}
% \end{flalign}
% \label{contact_eq}
% \end{subequations}
\begin{equation}
    \mathcal{Q}_\text{C} = \left\{ {z}^{i, p},  {\lambda}^{i, p}_n, \mathbf{p}^i \ \middle| \ 
    \begin{aligned}
        &\{{z}^{i, p} = 0\}   \Longrightarrow {\lambda}^{i, p}_n = {0},\\  
        &\sum_{p}^{N_p} {z}^{i, p} = 1,\\
        &\{{z}^{i, p} = 1\}   \Longrightarrow \mathbf{p}^i \in \mathcal{P}_p
    \end{aligned} 
    \right\}
\label{making_eq}
\end{equation}
% where $\lambda_n$ and $\lambda_f$ represent the normal and frictional components of contact forces. Similarly, $y_n$ and $y_f$ represent the normal and tangentional components of contact locations. $\mu$ is the corresponding friction constant. 
\eq{making_eq} means that 1) the normal force at $p$-th surface is zero if there is no contact, 2) the robot makes contact at one of the object surfaces, 
% including the case where ${\lambda}^{i, p}_n = {0}$
 and 3) the robot position is bounded in $\mathcal{P}_p$ if the contact is made. 
% $\mathbf{y}$ is the corresponding contact location (e.g., $\mathbf{p}^i, \mathbf{e}^j$). 
% $\mathcal{C}$ is the corresponding contact area which ensures that the contact is  

\subsubsection{Friction Cone}
We consider Coulomb friction model:
\begin{equation}
    \mathcal{Q}_\text{FC} = \{ \mathbf{f}^v, \boldsymbol{\lambda}^{i, p} |  | {\lambda}^{i, p}_s |\leq  \mu_r^i {\lambda}^{i, p}_n,  |{f}^{v}_s| \leq  \mu_e^v {f}^{v}_n \}
\end{equation}

% This model can be formulated as mixed-integer constraints using big-M formulation. 
\subsubsection{Stable Contact Change Constraint}
Although tabletop manipulation such as sliding \cite{graesdal2024convexrelaxations} can change contact anytime since the stability of the object is always maintained, the robot cannot change the contact anytime when working on non-tabletop manipulation such as pivoting since stability is not always maintained. 
% The work in \cite{zhang2023simultaneous} allows the change of contact in RRT-based planner if force and moment balance are still satisfied with removing robot contact, which can result in many iterations until RRT makes progress. 
In this work, \cont{} takes into account the change of contact stably.
% In this work, we consider this in \cont{} and mixed-integer optimizer can efficiently consider the change of the contact using the following simple mixed-integer linear constraints. 

The key idea is that the robot can safely change the contact when \eq{Mechanics} is satisfied with zero robot forces at $t$ and $t+1$ (i.e., ${\lambda}^{i, p}_{t} = {\lambda}^{i, p}_{t+1, n} = 0$). We can have three different scenarios.
\begin{enumerate}
    \item the robot does not make contact at $p$-th object surface between $t$ and $t+1$ (i.e., $z^{i, p}_t + z^{i, p}_{t+1} =0$).
    \item the robot changes the contact (i.e., $z^{i, p}_t + z^{i, p}_{t+1} =1$).
    \item the robot keeps making contact (i.e., $z^{i, p}_t + z^{i, p}_{t+1} =2$).
\end{enumerate}
Thus, we want to impose the constraint such that ${\lambda}^{i, p}_{t, n} = {\lambda}^{i, p}_{t+1, n} = 0$ when the first case happens, which can be implemented as mixed-integer linear constraints. 
\subsubsection{Sticking-Sliding Contact}
\quasi{} considers sticking-sliding contact.
% , which is not considered in \cont{}.
As a result, \quasi{} can find a much better trajectory than \cont{}.  
For each contact, $ \mathbf{f}^v, \boldsymbol{\lambda}^{i, p}, \dot{\mathbf{p}}_t^i, \dot{\mathbf{e}}_t^v$ are constrained through complementarity constraints to model sticking-sliding contact. See \cite{shirai2022robust, zhang2023simultaneous} for more details. 
% For each contact, 
% \begin{equation}
%     \mathcal{Q}_\text{slip} = \left\{ \mathbf{f}^v, \boldsymbol{\lambda}^{i, p}, \dot{\mathbf{p}}_t^i, \dot{\mathbf{e}}_t^v  \ \middle| \ 
%     \begin{aligned}
%         &0\leq \dot{y}^+_f \perp \mu \lambda_n - \lambda_f \geq 0\\  
%         &0\leq \dot{y}^-_f \perp \mu \lambda_n + \lambda_f \geq 0
%     \end{aligned} 
%     \right\}
% \label{stick_slide0}
% \end{equation}
% % \begin{subequations}
% % \begin{flalign}
% % 0\leq \dot{y}^+_f \perp \mu \lambda_n - \lambda_f \geq 0,\\
% % 0\leq \dot{y}^-_f \perp \mu \lambda_n + \lambda_f \geq 0
% % \end{flalign}
% % \end{subequations}
% where $\dot{y}_f = \dot{y}^+_f - \dot{y}^-_f$ and.... 
\section{Results}\label{sec:results}
In this section, we present numerical results for our proposed approach and compare them against some baselines. Through this section, we answer the following questions. 
\begin{enumerate}
    %\item How does our method work?
    \item Can the proposed framework discover complex multi-modal behavior?
    \item How long does it take to compute a solution for problems of different sizes? 
    \item Can the solutions found by the framework be implemented on real physical systems?
\end{enumerate}

\begin{figure}
    \centering
    \begin{subfigure}{0.2385\textwidth}
        \centering
        \includegraphics[width=\linewidth]{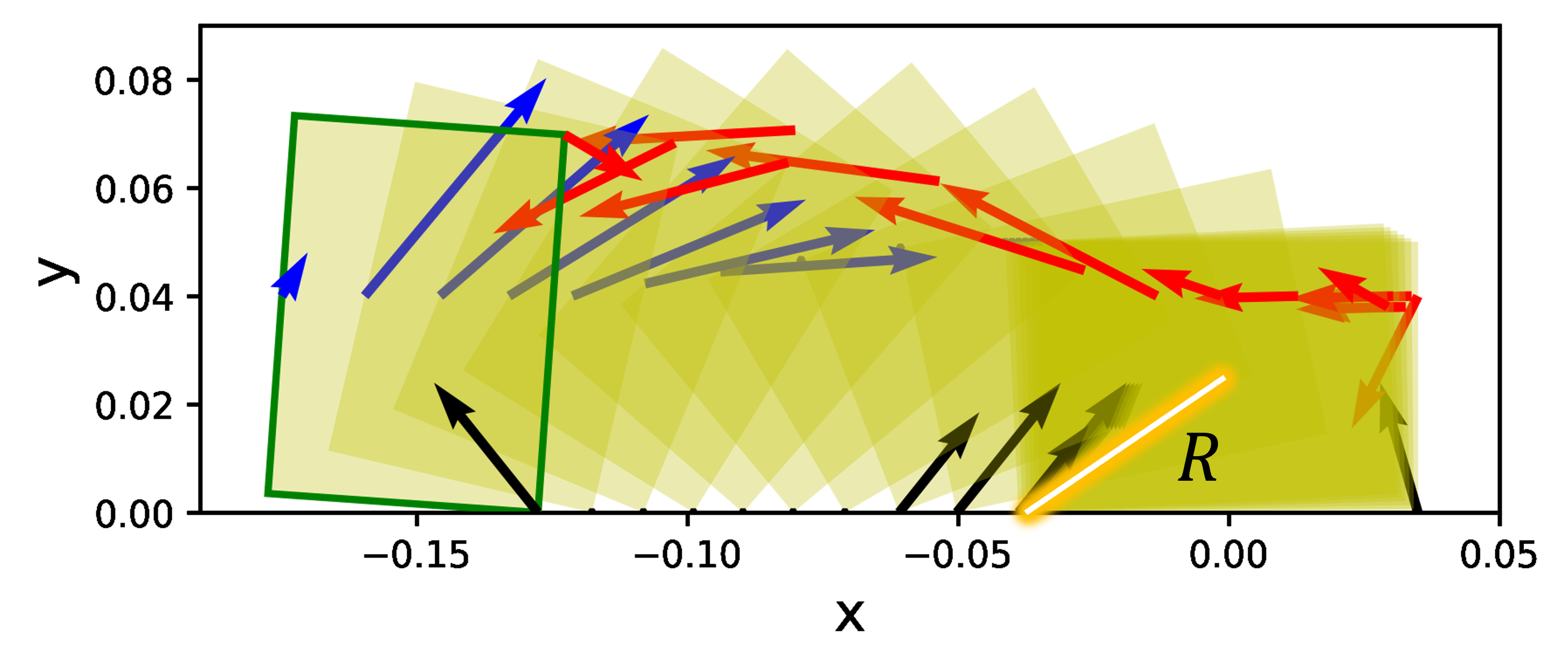}
        \caption{Without robustness cost.}
        \label{fig:pivot_non_robustness}
    \end{subfigure}
    \hfill
    \begin{subfigure}{0.2385\textwidth}
        \centering
        \includegraphics[width=\linewidth]{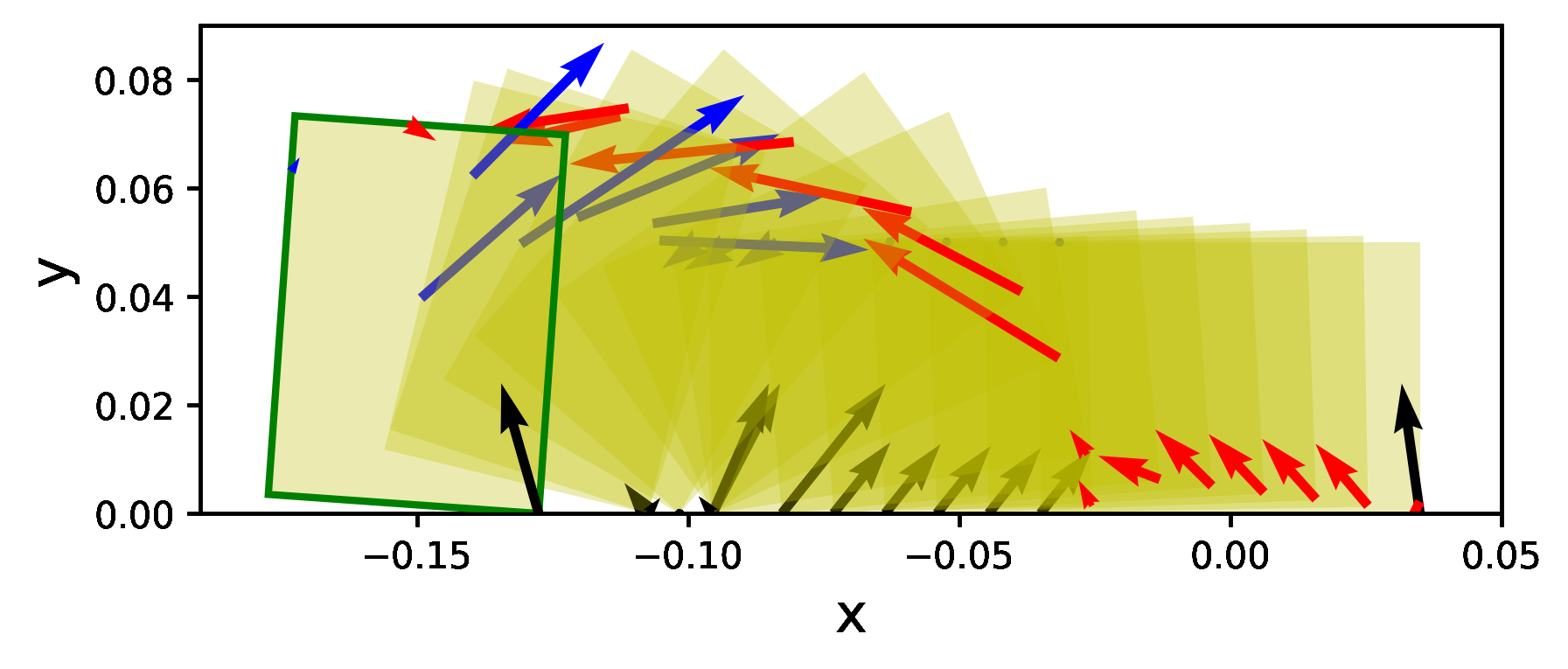}
\caption{With robustness cost.}
        \label{fig:pivot_robustness}
    \end{subfigure}
    \caption{Trajectories of pivoting manipulation from \quasi{}. We do not let one of the arms make contact if $q_y \geq \SI{-0.05}{\meter}$, denoted as blue arrows.  We consider $T=150$ but only plot 15 snapshots with no gravity arrows (aligned along $y$-axis.), for clarity. The bold green line represents the goal state. The orange line shows $R$ at $t = 0$ (see Sec~\ref{sec:main_result}).}
    \label{fig_pivot_stability}
\end{figure}

\subsection{Experiment Setup}
We implement our framework in Python using SNOPT \cite{gill2005snopt} with Drake \cite{drake} for solving \kine{} and \quasi{}.
We use Gurobi \cite{gurobi} for solving \cont{}. The computation is tested on a computer with Intel i9-13900K.  
% 
% As a baseline, given \kine{}, we solve the optimization problem ...
% We consider various objects in numerical evaluation and consider 2 different objects in hardware experiments.
For hardware experiments, we use a Mitsubishi Electric Assista arm with a stiffness controller, equipped with a WSG-32 gripper. The object pose is measured using AprilTag \cite{wang2016apriltag}.

\subsection{Results of Generated Trajectories using Our framework}\label{sec:main_result}
% show how multi-modal capability appears with sliding with contact switch

\begin{figure}
    \centering
    \begin{subfigure}{0.2385\textwidth}
        \centering
        \includegraphics[width=0.7\linewidth]{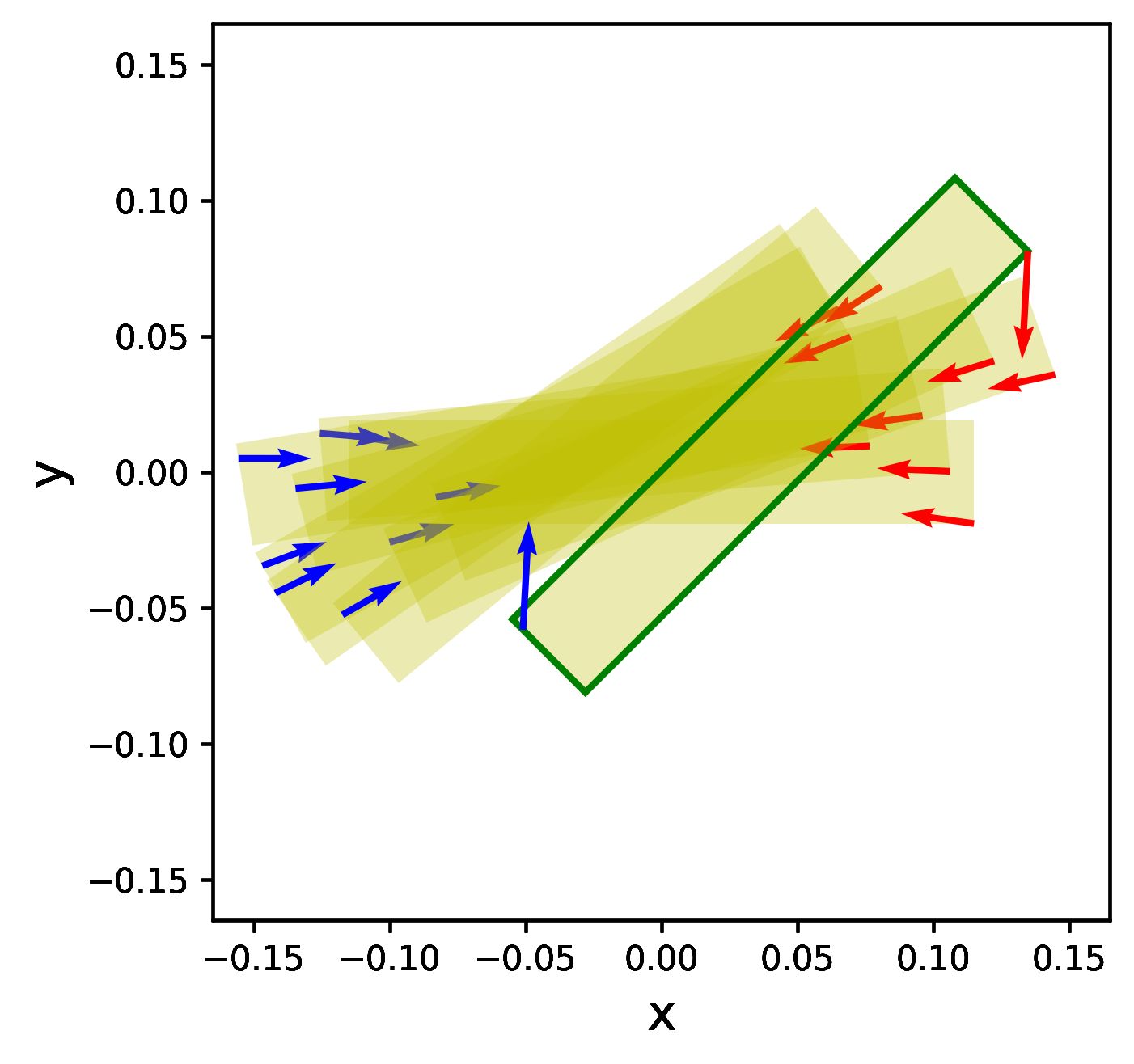}
        \caption{Output of \cont{}.}
        \label{fig:slip_cont_overlay}
    \end{subfigure}
    \hfill
    \begin{subfigure}{0.2385\textwidth}
        \centering
        \includegraphics[width=0.7\linewidth]{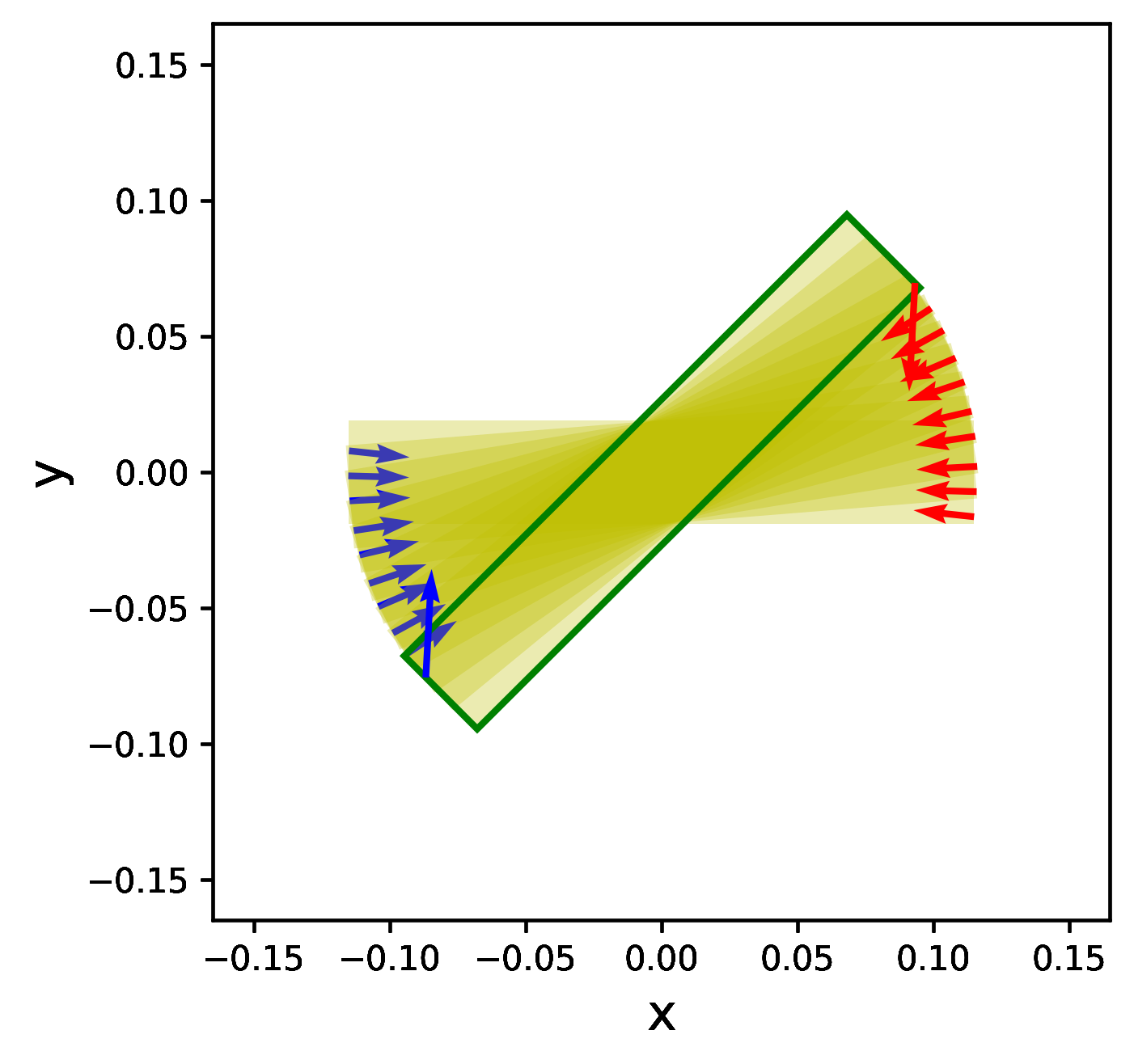}
\caption{Output of \quasi{}. }
        \label{fig:slip_quasi_overlay}
    \end{subfigure}
        \hfill
%     \begin{subfigure}{0.2385\textwidth}
%         \centering
%         \includegraphics[width=\linewidth]{figures/slip_quasi_overlay_one_arm.png}
% \caption{Output of \quasi{} with one arm.}
%         \label{fig:slip_quasi_overlay_one_arm}
%     \end{subfigure}
    \caption{Trajectories for sliding manipulation to rotate the object by \SI{45}{\degree}. Because \quasi{} can consider full nonlinear dynamics, it could achieve better motion than \cont{}. Note that we consider $T=50$ but only plot 10 snapshots for clarity. The bold green line represents the goal state. }
    \label{fig_quasi-vs-cont-rotation}
\end{figure}

\begin{figure}
    \centering
    \begin{subfigure}{0.2385\textwidth}
        \centering
        \includegraphics[width=0.6\linewidth]{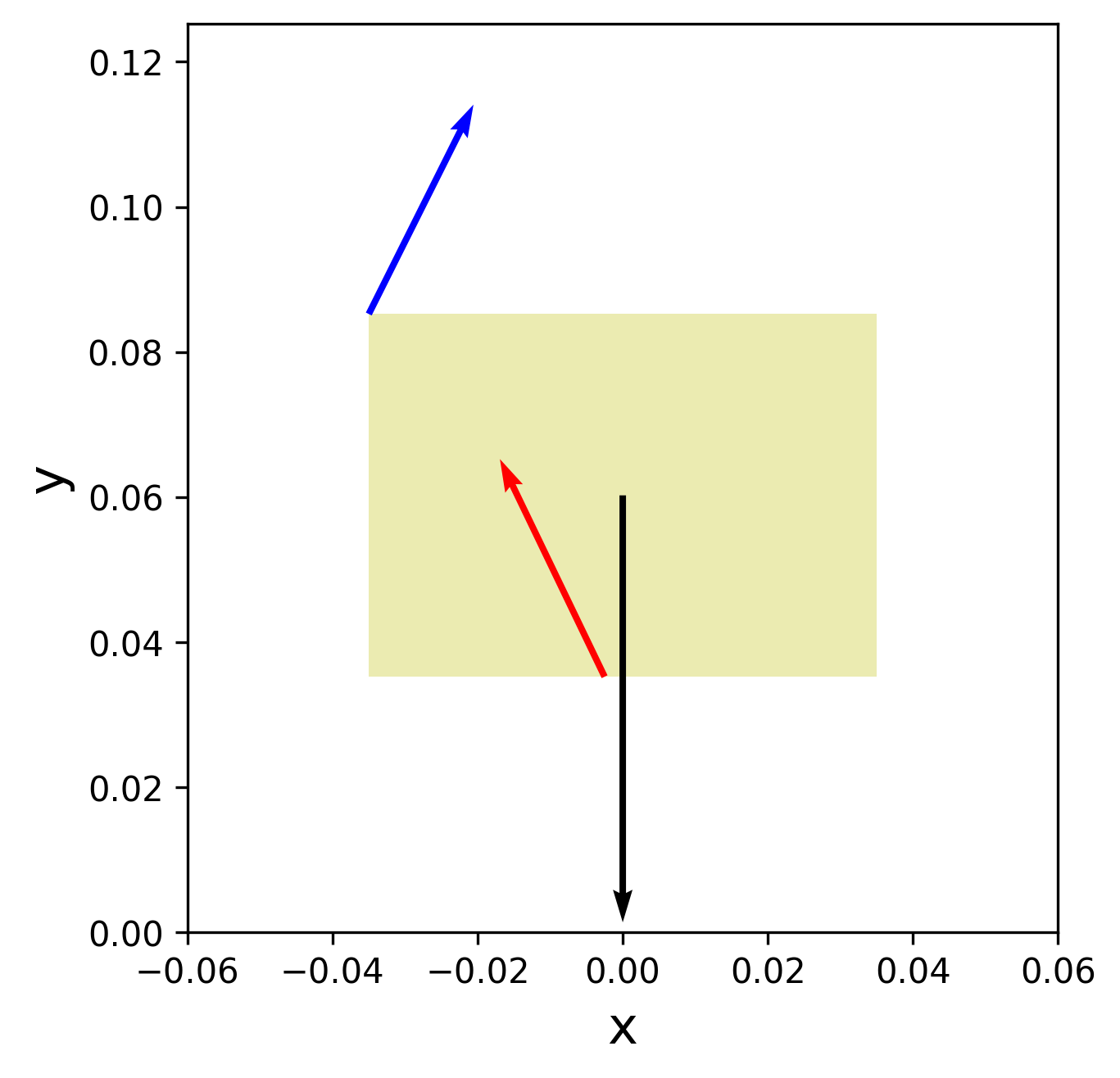}
        \caption{}
        \label{fig:no_envelope_overlay}
    \end{subfigure}
    \hfill
    \begin{subfigure}{0.2385\textwidth}
        \centering
        \includegraphics[width=0.6\linewidth]{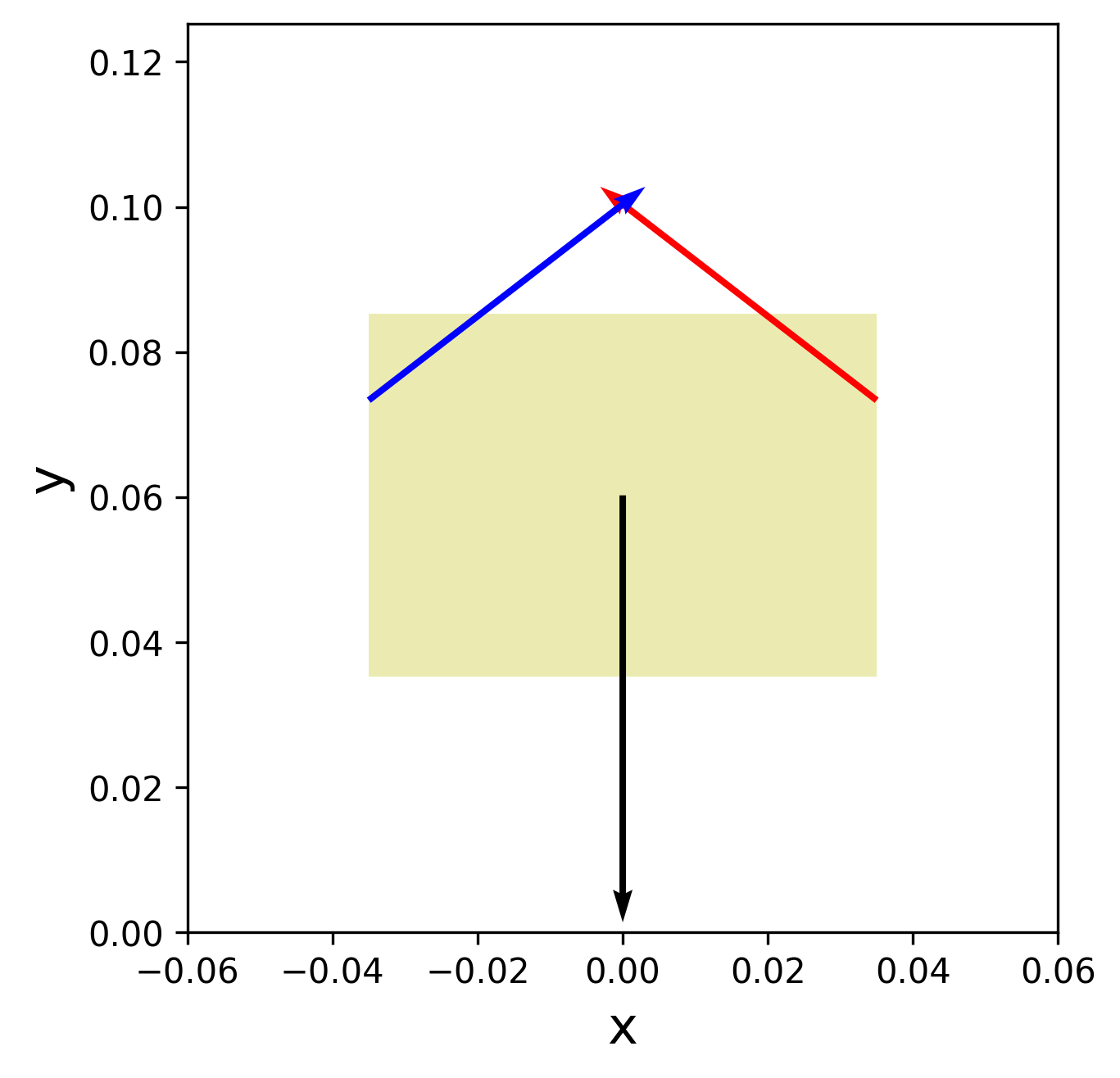}
\caption{}
        \label{fig:quad_envelope_overlay}
    \end{subfigure}
    \caption{We show how \cont{} generates robot contact location using (a) McCormick envelope and (b) ours. Black arrow shows gravity.
    % \ysnote{If necessary, I can also create plot showing the corresponding SQP result. I don't think it is necessary though.}
    }
    \label{fig_miqp_solution_change}
\end{figure}

\begin{figure*}[t]
    \centering
    \begin{subfigure}{0.495\textwidth}
        \centering
        \includegraphics[width=\linewidth]{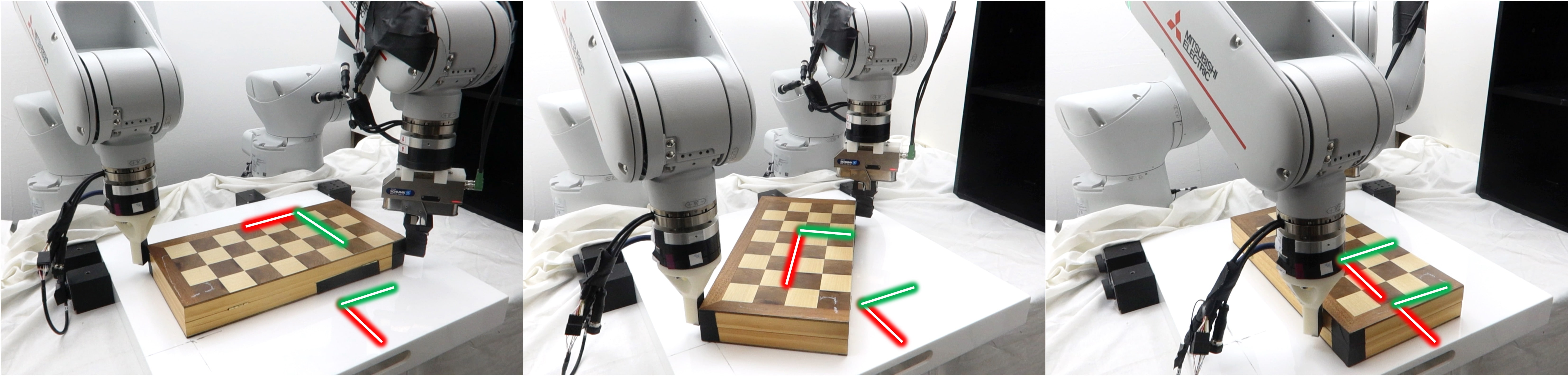}
        \caption{Dual arm sliding}
        \label{fig:bimanual-slide-hardware}
    \end{subfigure}
    \hfill
    \begin{subfigure}{0.495\textwidth}
        \centering
        \includegraphics[width=\linewidth]{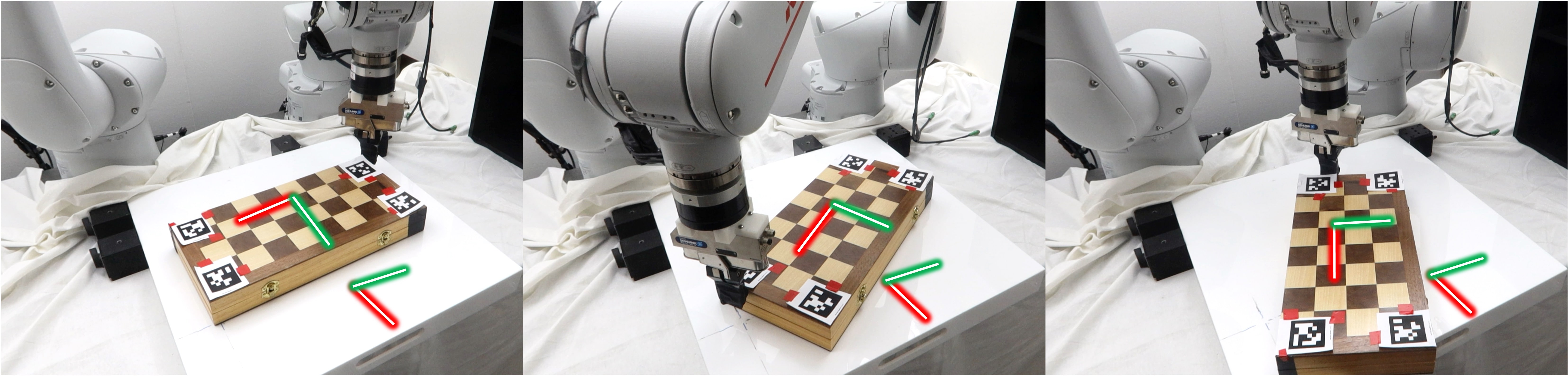}
\caption{Single arm sliding}
        \label{fig:single-arm-slide}
    \end{subfigure}
    \caption{Demonstration of sliding manipulation of a box in hardware experiments using the proposed algorithm. In \fig{fig:bimanual-slide-hardware}, the robots do not change the contact location on the object. In \fig{fig:single-arm-slide}, the robot has to change its contact location to accomplish its task. 
    The arrows represent the current orientation of the object and its desired orientation at the terminal time-step. 
    The Apriltag system is used for feedback during the change of the contact location to ensure that the robot makes contact in the correct location. 
    The hardware experiment video is found \href{https://youtu.be/s2S1Eg5RsRE?si=g7JP4_0Cchm49c2b}{here}.
    % \ysnote{If necessary, I can also create plot showing the corresponding SQP result. I don't think it is necessary though.}
    }
    \label{bimanual_vs-single-slide}
\end{figure*}

% \subsubsection{How much does \quasi{} polish the trajectories?}
First, we discuss how our method, \cont{} and \quasi{}, work together to achieve versatile contact-rich trajectories. To discuss this, we consider the pivoting task with one extrinsic contact with the environment. The goal is to rotate the object by \SI{90}{\degree}. We do not allow one of the arms to make contact if $q_y \geq \SI{-0.05}{\meter}$, to consider the workspace of the arm.  Given the trajectory from \cont{}, we consider two cases. In the 1st case, for each time step $t$, we consider the robustness cost, $\max (R (\cos{q_{\theta_t})^2})$ in \quasi{}, where $R$ is the distance between the extrinsic contact location and the center of mass of the object (see \fig{fig_pivot_stability}). In the 2nd case, we omit this cost. 

The results are shown in  \fig{fig_pivot_stability}. In \fig{fig:pivot_non_robustness}, the robot keeps pivoting the object until another arm starts supporting the object after $q_y \geq \SI{-0.05}{\meter}$. 
% Also, in \fig{fig:pivot_non_robustness}, the trajectory from \cont{} and the trajectory from \quasi{} are quite similar. 
In contrast, in \fig{fig:pivot_robustness}, one of the arms first pushes the object around $q_y = \SI{-0.05}{\meter}$ and then both arms pivot the object together. 
This is because using two arms minimizes the time the object remains unstable in gravity, ensuring better stability. Hence, that robust cost is maximized.
This is achieved because \quasi{} is capable of updating the trajectory of $\mathbf{q}_t$ based on user requirements. 
% Importantly, \cont{} does not need to provide 

Next, we discuss the scenario where the goal is to slide the object on the table by \SI{45}{\degree} with no position drifts. The results are shown in \fig{fig_quasi-vs-cont-rotation}. From \fig{fig:slip_cont_overlay} and \fig{fig:slip_quasi_overlay}, \quasi{} could achieve significantly less drift in position than \cont{}. This is because \quasi{} can consider the full nonlinear dynamics, and thus it can utilize the nonlinearity of the system. This highlights that our framework can provide a descent trajectory of the system as long as \cont{} provides decent contact information.
% As a result, \quasi{} can improve the optimality of the trajectory. 

% where \quasi{} achieves different multi-modal motion as illustrated in \fig{fig_pivot_stability}. 
% Using our framework, \quasi{} can actually change trajectory of the object pose and enables user to achieve more make sense trajectories. 

% (Also sliding also happens in \fig{fig_pivot_stability}, then we can argue that our framework can work even if MIQP does not give you super making sense solution. simple finger sliding)

% \subsubsection{How do the trajectories differ?}

% In conclusion, we confirm that our framework could successfully achieve because of its hierarchical structure. 

% \subsubsection{Can our framework design various trajectories?}
% ovNext, we show that our framework can design various motions as others can also do (\ysnote{show M (pushing), F (grasping), R (pulling), L (pivot)}. ).  
% Show sliding with 1 arm and 2 arm. Again emphasize that we do not choose where to make contact. 

% Also say that multi-modal capability can be also observed in \fig{fig_quasi-vs-cont-rotation}. Say again, we can observe how \quasi{} changes the trajectory and achieve more optimal trajectory.  

% what we want to show
% scenario 1: flat and show different motion with different physical parameter (e.g., friction) -> Consider multi-modal. Basically I need to show images 
% also say that contact is only switched when force is zero -> non-table top manipulation
% also show that how violation and contact force changes over time.

% show hardware experiment snapshots

\subsection{Results of Convex Relaxation of Bilinear Terms}
% show the 3 different results of MIQP.
% 1. Not considering any envelope
% 2. with linear envelope
% 3. with proposed one
% how much iteration improves

Next, we verify that the proposed convex relaxation of bilinear constraints results in tigher solutions. 
% We discuss how \cont{} changes its solution using different relaxations of bilinear terms. 
We consider a grasping task where the goal is to lift the object. We consider \cont{} using McCormick envelope \eq{McCormick} and ours \eq{binary_encode} with $K=2$.
The results are illustrated in \fig{fig_miqp_solution_change}. In \fig{fig:no_envelope_overlay}, it is obvious that there is clock-wise moment exists while in \fig{fig:quad_envelope_overlay}, moment balance is satisfied. Thus, our proposed relaxation could achieve a tighter estimation of bilinear terms. 
We also observe that our method where \cont{} uses \eq{binary_encode} shows a higher success rate than our method where \cont{} uses \eq{McCormick} (see \tab{tab:computation}) with some cost of computation.
This makes sense since \quasi{} has higher chances of finding feasible solutions if \cont{} provides better solutions. 
% Therefore, we argue that using our relaxation method can be more beneficial if finding feasible solutions with a higher chance is preferred.
% , and using the McCormick envelope might make more sense if the computation is prioritized. 

\subsection{Computational Results}
% argue that this works well
% success rate of each task
% runtime with variance for the results when it finds solutions
% show the results with longer time horizon
% against gurobi and binary using. Also NYU one if I have time
% show how many violations
We discuss the computational results. We consider 500 different initial and goal states of the box whose width and height are \SI{0.07}{\meter} and \SI{0.05}{\meter}, respectively. 
We sample initial and goal states of the object uniformly from the range of $\pm\SI{0.05}{\meter}$ of $x$- and $y$-positions, and $\pm\SI{90}{\degree}$ of the angle of the object, respectively.
For each initial and goal pose of the object, we run optimizers. For all samples, we use $h = 0.2$s. 
% For the comparison, we use \gurobi{} \cite{gurobi} as a baseline to solve the original mixed-integer non-convex QCQP \eq{cont_eq} without using any decomposition of the optimization problem and any relaxation for bilinear constraints we present in this paper.
All options are summarized in \tab{tab:computation}.

\textbf{Success Rate.}
We consider $T = 200$. 
We define success as the optimizer finding a feasible solution within 1 minute. We denote $S_i$ as the success rate of the $i$-th option and we got $S_1 = 8 \%, S_2 = 22 \%, S_3 = 41 \%, S_4 = 71 \%$.
% Thus, we verify that our method could successfully find solutions within the time frame over various different problems. 

\textbf{Computation Time.}\label{sec:compt_time}
The results are summarized in \tab{tab:computation}. 
Overall, we observe that option 4 (ours) shows the best performance. 
First, the total computation time in option 4 is generally the smallest among options. 
Second, option 4 and option 1 show smaller $a_2$ (defined in \tab{tab:computation}) than other options. This makes sense since for these options, \cont{} converges quickly because of numerical efficiency. 
Third, we observe that the number of iterations in option 2, to add cutting plane denoted as $a_3$, is largest compared to other methods because the naive McCormick envelope does not provide good quality solutions due to relatively rough approximation of bilinear constraints, resulting in many cuts. 

\subsection{Hardware Experiments}
We implement the trajectories generated by our framework on a robotic system in open-loop. We consider two tasks, a stowing task where the goal is to stow an object on a shelf, and a sliding task where the goal is to rotate the object by \SI{90}{\degree}. The results are shown in \fig{fig1} and \fig{bimanual_vs-single-slide}. We observe that the robot(s) could successfully achieve the desired tasks in \fig{fig1} and \fig{fig:bimanual-slide-hardware}. Note that  \fig{fig:single-arm-slide} shows that the single arm could not complete its task due to open-loop control.

\begin{table}[]
    \centering
    \caption{Computation results for different options and time horizons. 
    Each option differs in how \cont{} computes. 
    Option 1 is the baseline where \gurobi{} solves the original mixed-integer non-convex QCQP \eq{cont_eq} in \cont{}. Option 2 uses the naive McCormick envelope \eq{McCormick}.
    Option 3 uses the piecewise McCormick envelope \eq{general_relaxation} with $C=8$.
    Option 4 uses our piecewise McCormick envelope using binary encoding \eq{binary_encode} with $K=3$. 
    Each entry contains three values: \(a_1, a_2, a_3\).
    $a_1$ is the average total computation time (i.e., runtime of \cont{} + \quasi{}) in second. 
    $a_2$ is the average ratio of the computation time spent in \cont{} over $a_1$. 
    $a_3$ is the average number of cuts (i.e., the number of iterations between \cont{} and \quasi{}). 
    We compute these values over the successful samples defined in Sec.~\ref{sec:compt_time}.
    }
    \setlength{\tabcolsep}{6pt} % Adjust column spacing here
    \begin{tabular}{|c|c|c|c|}
        \hline
        \multirow{2}{*}{Options} & \multicolumn{3}{c|}{Time Horizon ($T$)} \\
         & $T = 50$ & $T = 100$ & $T = 200$   \\
        \hline
        1  & (28, 0.80, 1.0) & (56, 0.79, 1.9) & (59, 0.89, 2.1)  \\
        2  & (0.9, 0.14, 18.5) & (50, 0.11, 25.5) & (59, 0.29, 70.2)  \\
        3  & (10, 0.53, 7.1) & (44, 0.80, 10.9) & (55, 0.88, 21.9)  \\
        4  & (6, 0.30, 6.1) & (9, 0.35, 10.1) & (11, 0.55, 20.1) \\
        % 5  & (x13, y13, z13) & (x14, y14, z14) & (x15, y15, z15)  \\
        \hline
    \end{tabular}
    \label{tab:computation}
\end{table}

% Box pivoting and box sliding with 2 arms and 1 arm.
We discuss the results of the sliding task. 
% We discuss the results from \fig{fig:bimanual-slide-hardware} and \fig{fig:single-arm-slide}. 
In \fig{fig:bimanual-slide-hardware}, we consider two arms while in \fig{fig:single-arm-slide} we consider a single-arm manipulation. We observe that in \fig{fig:bimanual-slide-hardware}, the bimanual maintains the same contact location but in \fig{fig:single-arm-slide} the single arm keeps changing the contact location on the object so as to achieve the desired goal. Therefore, our framework can automatically achieve very different manipulation behavior given underlying constraints (in this case, single point contact vs two contacts). These experiments also verify that our algorithm can generate physically meaningful solutions. %since our method can consider the object, robots, and contact simultaneously, and we verify its multi-modal capability in the hardware experiments.

% \subsection{Results of Relaxation of Bilinear Terms}

% (if have time), different 3 different objects such as triangle and "L shape"

% We discuss the computational results in this section. 

% Show the following results / plots:

\section{Discussion}\label{sec:discussion}
In this paper, we propose a framework for efficiently computing optimal trajectories of an object, robots, and contact simultaneously using hierarchical optimization.
% Our method is able to solve general complex manipulation tasks that are not intu
% various complex constraints, which are often missing in the literature, such as changing contact modes safely, sticking-sliding complementarity constraints, and force $\&$ moment balance constraints with extrinsic contacts. 
% Our method decomposes the original MINLP into a smaller optimization problem, where each optimization problem can find feasible solutions computationally efficiently.
We decompose the original problem into \cont{} and \quasi{} to handle multi-modal manipulation tasks efficiently.
In particular, we present a convex relaxation of bilinear constraints using the binary encoding method so that \cont{} can provide tighter solutions.  
% This was accomplished by our decomposition optimization, novel relaxation of bilinear terms, and feedback cutting plane. 
We demonstrate and verify our method extensively under various scenarios including hardware experiments. 

\textbf{Future Work.}
% Our work has several limitations which we would like to investigate in the future.
Although our method can realize multi-modal manipulation, our method may fail if the trajectory by \kine{} does not satisfy system dynamics. Thus, the key question we want to answer is how we obtain a kinematics trajectory in \kine{} that has a high chance of satisfying the system dynamics. Ultimately, we want to understand how to update the kinematics trajectory when the proposed method fails to find a feasible solution based on the infeasible solution.
Similarly, we are interested in working on adding more informative cuts to \cont{} if \quasi{} only finds infeasible solutions. Our current cutting-plane method does not fully utilize the infeasible solution by \quasi{} so it is interesting how to encode informative information to add better cuts.
Lastly, this paper focuses on contact at the robot’s end-effector, but we aim to extend our approach to whole-body robot contacts (e.g., \cite{murooka2015whole, pang2023global, shirai2024linear, leve2024explicit}) for more generalized multi-modal manipulation.

% Another natural extension of this work is the manipulation in 3D. 
% \textbf{Manipulation in 3D}: 
% The natural extension of this work is the manipulation in 3D, which increases the complexity of the problem dramatically. To achieve this, we are interested in formulating the optimization problem using decomposition-based optimization techniques to find the solution quickly. 

% \textbf{Utilization of non-good solution}: 
% Although we remove the contact combination from \cont{} if \quasi{} does not converge, 

% In this work, we have not analyzed some properties of our method formally, such as convergence analysis. 

% \textbf{Feedback in Kinematics Trajectory Optimization}:
% Although our method can realize multi-modal manipulation tasks, our method might not work if the trajectory by \kine{} does not satisfy dynamics of the system. Thus, the key question we would like to answer is how we get kinematics trajectory in \kine{} that has high chance of satisfying the dynamics of the system.
% Additionally, we are interested in how to make our method consider the collision between the robot and the object, or even the object and another object, to achieve more general manipulation tasks.

% Users need to specify waypoints. \ysnote{maybe too much details}

%%%%%%%%%%%%%%%%%%%%%%%%%%%%%%%%%%%%%%%%%%%%%%%%%%%%%%%%%%%%%%%%%%%%%%%%%%%%%%%%

\bibliographystyle{IEEEtran}
\bibliography{main}

\end{document}